\documentclass[lettersize,journal]{IEEEtran}
\usepackage{amsmath,amsfonts}
\usepackage{algorithm}
\usepackage{array}
\usepackage{textcomp}
\usepackage{stfloats}
\usepackage{url}
\usepackage{verbatim}
\usepackage{graphicx}
\usepackage{cite}
\usepackage[pagebackref=true,colorlinks,bookmarks=true,citecolor=green,linkcolor=blue,urlcolor=blue]{hyperref}

\usepackage{url}            
\usepackage{booktabs}       
\usepackage{amsfonts}       
\usepackage{nicefrac}       
\usepackage{microtype}      
\usepackage{graphicx}
\usepackage{makecell}
\usepackage{multirow}
\usepackage{bm}
\usepackage{booktabs} 
\usepackage{overpic}
\usepackage{pict2e}
\usepackage{amsmath}
\usepackage[table]{xcolor}

\usepackage{rotating}
\usepackage{amssymb}
\usepackage{pifont}
\usepackage{caption}
\usepackage{subcaption}
\usepackage{ragged2e}

\usepackage{setspace}
\usepackage{listings}
\usepackage{anyfontsize}
\usepackage{colortbl}

\usepackage[capitalize]{cleveref}
\crefname{section}{Sec.}{Secs.}
\Crefname{section}{Section}{Sections}
\Crefname{table}{Table}{Tables}
\crefname{table}{Tab.}{Tabs.}

\definecolor{myblue}{RGB}{169,196,235}
\definecolor{mygreen}{RGB}{213,232,212}
\definecolor{mygray}{RGB}{191,191,191}
\definecolor{mycolor}{RGB}{184,96,41}

\newcommand{\myPara}[1]{\vskip 0.05in\noindent\textbf{#1}}

\ifdefined \GramaCheck
\newcommand{\CheckRmv}[1]{}
\newcommand{\figref}[1]{Figure 1}%
\newcommand{\tabref}[1]{Table 1}%
\newcommand{\secref}[1]{Section 1}
\renewcommand{\eqref}[1]{Equation 1}
\else
\newcommand{\CheckRmv}[1]{#1}
\newcommand{\figref}[1]{Fig.~\ref{#1}}%
\newcommand{\tabref}[1]{Tab.~\ref{#1}}%
\newcommand{\secref}[1]{Sec.~\ref{#1}}
\newcommand{\appref}[1]{App.~\ref{#1}}
\renewcommand{\eqref}[1]{Eqn.~(\ref{#1})}
\fi
\def\vs{\emph{vs.}}

\begin{document}

\title{Mr. DETR++: Instructive Multi-Route Training for Detection Transformers with Mixture-of-Experts}

\author{Chang-Bin Zhang, Yujie Zhong and Kai Han
\thanks{Chang-Bin Zhang (cbzhang@connect.hku.hk) and Kai Han (kaihanx@hku.hk) are with The University of Hong Kong; Yujie Zhong (jaszhong@hotmail.com) is with Meituan Inc.}
\thanks{Kai Han is the corresponding author.}
}

\markboth{}%
{Shell \MakeLowercase{\textit{et al.}}: A Sample Article Using IEEEtran.cls for IEEE Journals}


\maketitle

\begin{abstract}
Existing methods enhance the training of detection transformers by incorporating an auxiliary one-to-many assignment.
In this work, we treat the model as a multi-task framework, simultaneously performing one-to-one and one-to-many predictions.
We investigate the roles of each component in the transformer decoder across these two training targets, including self-attention, cross-attention, and feed-forward network.
Our empirical results demonstrate that any independent component in the decoder can effectively learn both targets simultaneously, even when other components are shared.
This finding leads us to propose a multi-route training mechanism, featuring a primary route for one-to-one prediction and two auxiliary training routes for one-to-many prediction.
We propose a novel instructive self-attention mechanism, integrated into the first auxiliary route, which dynamically and flexibly guides object queries for one-to-many prediction.
For the second auxiliary route, we introduce a route-aware Mixture-of-Experts (MoE) to facilitate knowledge sharing while mitigating potential conflicts between routes.
Additionally, we apply an MoE to low-scale features in the encoder, optimizing the balance between efficiency and effectiveness.
The auxiliary routes are discarded during inference.
We conduct extensive experiments across various object detection baselines, achieving consistent improvements as demonstrated in~\figref{fig:improvements}.
Our method is highly flexible and can be readily adapted to other tasks. To demonstrate its versatility, we conduct experiments on both instance segmentation and panoptic segmentation, further validating its effectiveness. Project page: \url{https://visual-ai.github.io/mrdetr/}
\end{abstract}
\begin{IEEEkeywords}
Object Detection, detection transformers, multi-route training, instructive self-attention, mixture-of-experts.
\end{IEEEkeywords}

\section{Introduction}\label{sec:introduction} 
\begin{figure}[htp]
    \centering
    \begin{overpic}[width=1\linewidth]{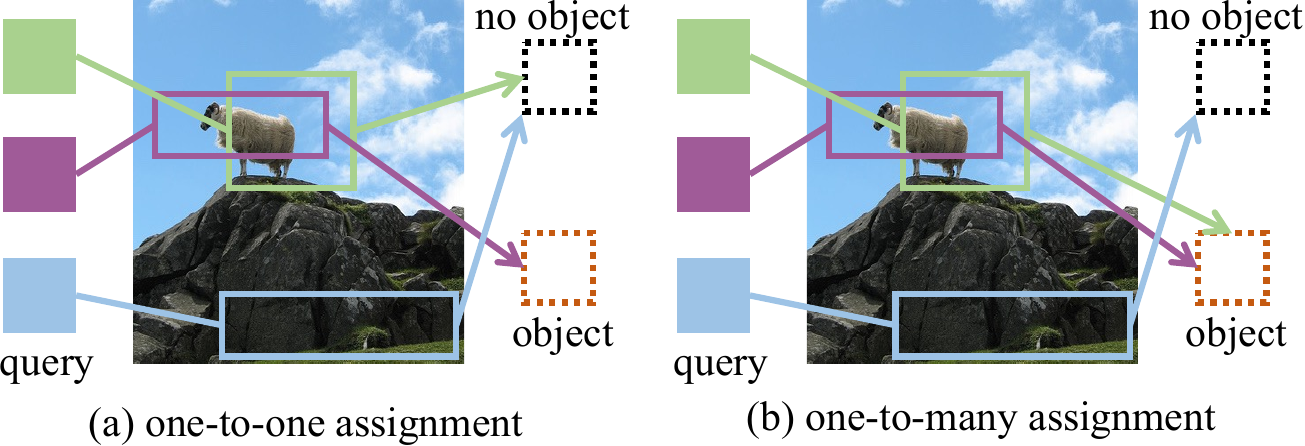}
    \end{overpic}
    \caption{\textbf{One-to-one \vs~one-to-many assignments.}
    (a) `One-to-one' assigns each ground-truth box to \textit{a single} predicted box, while predictions not associated with any object are supervised by `no object' (background).
    (b) `One-to-many' allows each ground-truth box to be paired with \textit{multiple} predicted boxes.
    }\label{fig:assignments}
    \label{fig:enter-label}
    \vskip -0.1in
\end{figure}

\IEEEPARstart{T}he end-to-end detection transformer (DETR)~\cite{detr}, along with subsequent studies~\cite{liu2022dab,meng2021conditional,wang2022anchor}, shine in object detection due to its simplicity and effectiveness.
Unlike traditional methods, DETR-based detectors eliminate the need for post-processing steps like non-maximum-suppression (NMS).
They employ the one-to-one assignment strategy (see~\figref{fig:assignments}(a)) for supervision, where each ground-truth box is matched to a single prediction.
Compared to the one-to-many assignment methods (see~\figref{fig:assignments}(b)) in conventional object detectors~\cite{ren2016faster,ross2017focal,tian2020fcos} that allow each ground-truth box to be assigned with multiple predictions, the one-to-one assignment can lead to slow convergence~\cite{hdetr,dino,sun2021rethinking} of DETRs due to its sparse supervision.

\begin{figure}
    \centering
    \includegraphics[width=1\linewidth]{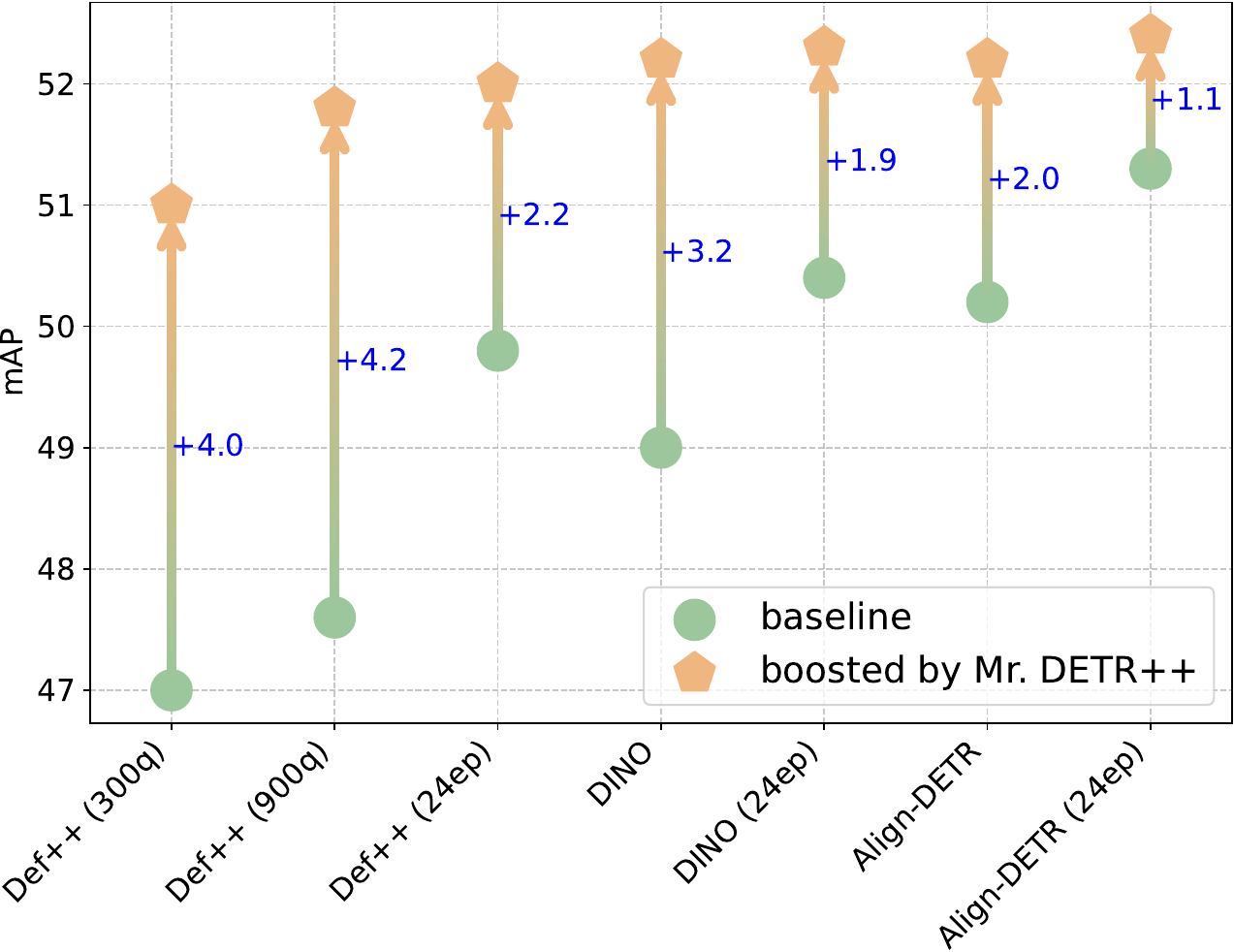}
    \caption{\textbf{Improvements over different baseline models.} The results are reported on the COCO 2017 validation in mean Average Precision (mAP).
        From left to right: Deformable-DETR++~\cite{deformable} with 300 queries, Deformable-DETR++~\cite{deformable} with 900 queries, Deformable-DETR++~\cite{deformable} trained for 24 epochs, DINO~\cite{dino}, DINO trained for 24 epochs, Align-DETR~\cite{cai2023align} and Align-DETR~\cite{cai2023align} configured with 24 training epochs.
    }
    \label{fig:improvements}
\end{figure}

\begin{table}[!thp]
\centering
\setlength{\tabcolsep}{3pt} 
\caption{\textbf{The AP performance of different variants.} Each variant includes a primary route for one-to-one prediction and several auxiliary routes for one-to-many prediction. `o2o': the performance of the primary route. `o2m': the maximum performance with NMS among the auxiliary routes. }\label{tab:motivation}
\resizebox{0.45\textwidth}{!}{
\begin{tabular}{c|l|c|cc}
\toprule
No & Configurations & Routes &  o2o & o2m\\
\midrule
(1) & One-to-one only & 1 & 47.6 & - \\
(2) & Share All & 1 & 41.6 (\textcolor{red}{-6.0}) & 41.6 \\
\midrule
(3) & Not shared Self-Attention & 2 & 49.7 (\textcolor{blue}{+2.1}) &  50.3 \\
(4) & Not shared Cross-Attention & 2 & 49.2 (\textcolor{blue}{+1.6})  & 50.0 \\
(5) & Not shared FFN & 2 & 49.6 (\textcolor{blue}{+2.0})  &  50.1 \\
\midrule
(6) & Shared Self-Attention & 2 & 49.4 (\textcolor{blue}{+1.8}) & 50.3 \\
(7) & Shared Cross-Attention & 2 & 49.4 (\textcolor{blue}{+1.8}) & 50.0 \\
(8) & Shared FFN & 2 & 49.2 (\textcolor{blue}{+1.6}) & 50.0 \\
\midrule
(9) & (3) + (4)  & 3 & 49.4 (\textcolor{blue}{+1.8}) & 49.9 \\
(10) & (3) + (5) & 3 & \textbf{50.0} (\textcolor{blue}{+2.4}) & \textbf{50.8} \\
(11) & (4) + (5) & 3 & 49.0 (\textcolor{blue}{+1.4}) & 49.6 \\
(12) & (3) + (4) + (5) & 4 & 49.6 (\textcolor{blue}{+2.0}) & 50.2  \\
\bottomrule
\end{tabular}
}

\vskip -0.1in
\end{table}

\begin{figure*}[t]
    \centering
    \begin{overpic}[width=\linewidth]{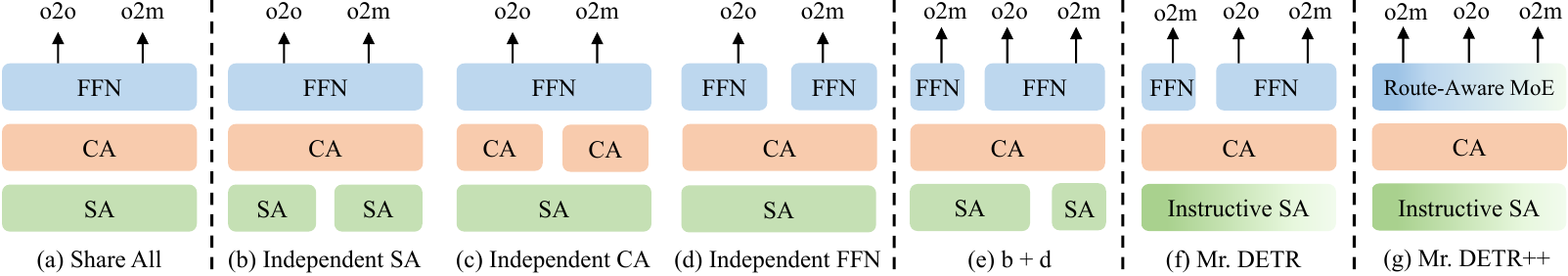}
    \end{overpic}
    \caption{\textbf{Different configurations of the transformer decoder with auxiliary one-to-many training.} `SA': self-attention. `CA': cross-attention. `FFN': feed-forward network. `o2o': one-to-one prediction. `o2m': one-to-many prediction.}\label{fig:motivationfig}
    \vskip -0.1in
\end{figure*}

To accelerate the training of DETR-like object detectors, several works propose auxiliary training methods to improve the quality of prediction localization by introducing auxiliary one-to-many assignment~\cite{hdetr,hu2024dac,msdetr} or multiple groups of one-to-one assignment~\cite{li2022dn,dino,chen2023group}.
Specifically, DN-DETR~\cite{li2022dn}, Group-DETR, and DINO~\cite{dino} utilize multiple groups of parallel axillary queries that share the same transformer decoder with the primary object query.
DETA~\cite{ouyang2022nms} finds that self-attention is necessary when the model is required to conduct one-to-one prediction, but not when the model performs one-to-many prediction.
Based on this observation, DAC-DETR~\cite{hu2024dac} and MS-DETR~\cite{msdetr} incorporate one-to-many auxiliary training by explicitly constraining self-attention and cross-attention to make the one-to-one and one-to-many prediction, respectively.
We regard the model architecture that performs one-to-one and one-to-many prediction simultaneously as a multi-task framework.
However, previous works examine the functions of each component in the transformer decoder for two training targets only in the single-task setting.
A similar investigation in a multi-task framework still needs to be conducted in a rigorous way.

In this work, we first build the multi-task framework and empirically investigate the roles of each component in the transformer decoder for one-to-one and one-to-many assignments, including self-attention, cross-attention, and the feed-forward network (FFN).
Specifically, we regard the model with auxiliary training by one-to-many assignment as a typical multi-task framework, requiring simultaneous one-to-one and one-to-many predictions.
As illustrated in~\figref{fig:motivationfig}(a), the most straightforward approach is to obtain the prediction results of two tasks with shared components.
However, experimental results in~\tabref{tab:motivation}(1) indicate that \emph{incorporating a one-to-many assignment significantly degrades the performance of the primary one-to-one prediction when all components are shared between two tasks.}
Consequently, we investigate which component can be independent to enhance the primary route's performance.
 We build the multi-task framework with independent self-attention, cross-attention, and FFN for two training targets, as shown in~\figref{fig:motivationfig}(b-d), respectively.
We find that \emph{any independent component significantly benefits the primary route of one-to-one prediction, even when other components are shared}, achieving 2.1\%, 1.6\%, and 2.0\% improvement, respectively, as demonstrated in~\tabref{tab:motivation}(3-5).
We also explore the different auxiliary routes with two independent components, which show similar performance to those with a single independent component, but involve more trainable parameters.
Therefore, we do not consider auxiliary training routes with two independent components.

This empirical finding motivates us to build a multi-route training mechanism that combines multiple auxiliary training routes, each with an independent component.
Specifically, the primary route is used for the one-to-one prediction, while each auxiliary route features a distinct independent component for one-to-many prediction.
For instance, in~\figref{fig:motivationfig}(e), we integrate auxiliary training using independent self-attention and FFN, respectively, with the primary route.
We conduct experiments to verify the performance of each combination in~\tabref{tab:motivation}.
We find that when involving an auxiliary route with independent cross-attention degrades the primary route's performance, likely due to the slow convergence~\cite{sun2021rethinking} of independent cross-attention.
The combination in~\figref{fig:motivationfig}(e) outperforms other variants, leading us to adopt this framework as our potential solution.

To reduce additional trainable parameters in the auxiliary training routes and promote parameter sharing across all routes, we propose a novel instructive self-attention mechanism, depicted in~\figref{fig:motivationfig}(f) named Mr. DETR.
It includes three training routes: a primary route for one-to-one prediction and two auxiliary routes, incorporating instructive self-attention and an independent FFN, respectively.
The first auxiliary route shares all parameters with the self-attention in the primary route, but incorporates a learnable token, named instruction token, attached to the input object queries in the self-attention.
The object queries, along with the instruction token, undergo self-attention, enabling the instruction token to guide the queries for one-to-many prediction.
We directly employ the independent FFN in the second auxiliary route.
During inference, the two auxiliary training routes are discarded, ensuring the model architecture and inference time remain consistent with baseline models.

To share knowledge between two independent FFNs designed as distinct experts for different prediction targets, we further propose a route-aware mixture-of-experts (MoE) framework, with which Mr. DETR evolves into Mr. DETR++, by replacing the two independent FFNs with the proposed route-aware MoE (see~\figref{fig:motivationfig}(g)).
Mr. DETR++ incorporates three routes: a primary route for one-to-one prediction and two auxiliary routes for one-to-many prediction.
In this architecture, all routes share a group of experts, with the primary route and one auxiliary route (employing instructive self-attention) sharing the same gating mechanism, while another auxiliary route uses independent gating to mitigate conflicts between routes.
This design enables the route-aware MoE to share knowledge across experts while avoiding conflicts between routes.
To extend the MoE framework to the transformer encoder, we introduce a scale-aware MoE strategy to address the computational cost of processing long sequences formed by flattened multi-scale image features. 
This strategy shares an expert across all tokens and disables gated experts for higher scales, significantly reducing computational overhead. 
Additionally, we propose a localization-aware calibrating strategy that, beyond the vanilla classification score, learns an IoU-related score during training, inspired by VarifocalNet~\cite{zhang2021varifocalnet}, to jointly assess classification and localization quality.
Unlike Stable-DINO~\cite{liu2023detection} and Rank-DINO~\cite{pu2024rank}, this score is excluded from the label assignment process, enabling it to serve as a simple plug-in module. 
During inference, the localization-aware score calibrates the classification score, enhancing prediction accuracy.

{\color{black}
In summary, we make the following main contributions:
\begin{itemize}
    \item We demonstrate empirically that, within a multi-task framework, any independent component in the decoder can effectively learn both one-to-one and one-to-many prediction targets simultaneously, even when other components are shared. 
    \item Leveraging this insight, we propose a multi-route training mechanism enhanced by a novel instructive self-attention mechanism that dynamically guides object queries for one-to-many prediction, culminating in the development of Mr. DETR.
    \item We introduce a route-aware mixture-of-experts (MoE) framework to enable knowledge sharing between the two independent feed-forward networks (FFNs) in Mr. DETR. By further incorporating a scale-aware MoE in the transformer encoder, we develop Mr. DETR++.
    \item We investigate the intrinsic mechanisms of our proposed multi-route training framework through probing experiments. Our findings reveal that auxiliary training routes mitigate conflicts with the primary route by selectively disrupting information necessary for one-to-one prediction within object queries.
    \item We validate the effectiveness of our approach through extensive experiments on multiple benchmarks, including COCO 2017, Objects365, and NuImages, achieving consistent improvements over various baseline models. Additionally, we extend our method to diverse tasks, such as instance segmentation and panoptic segmentation.
\end{itemize}
}

{\color{black}
A preliminary version of this work has been presented in~\cite{zhang2025mr}, which investigates the role of transformer decoder components in a multi-task framework and introduces a multi-route training mechanism with a novel instructive self-attention mechanism. 
In this work, we further propose a route-aware MoE framework to enable knowledge sharing while mitigating conflicts between routes for one-to-one and one-to-many predictions. 
Additionally, we introduce a scale-aware MoE in the transformer encoder to enhance feature extraction capabilities. 
By incorporating a localization-aware score, we develop a simple yet effective score calibration method. 
To demonstrate the effectiveness of our approach, we provide extensive experimental results on additional benchmarks, including Objects365~\cite{shao2019objects365} and NuImages~\cite{caesar2020nuscenes}, and extend our method to the panoptic segmentation task.
Furthermore, we conduct probing experiments to thoroughly investigate the intrinsic mechanisms of our proposed multi-route training framework.
}

\section{Related Work}

\subsection{Object Detection with Transformers}
Unlike traditional CNN-based object detectors~\cite{ren2016faster,zheng2021yolox,tian2020fcos,jin2022you}, DETR~\cite{detr} achieves end-to-end object detection without requiring any post-processing, leveraging a query-based transformer and set prediction mechanism.
To enhance performance and accelerate the training of DETR~\cite{detr}, numerous follow-up works~\cite{dai2021dynamic,gao2021fast,deformable} introduce advanced attention architectures. 
Specifically, Deformable-DETR~\cite{deformable} introduces deformable attention to enable efficient multi-scale attention.
Cascade-DETR~\cite{ye2023cascade} improves the performance of deformable attention by utilizing predicted boxes.
EASE-DETR~\cite{gao2024ease} enhances one-to-one prediction by modulating self-attention with the location relationship of queries, while Relation-DETR~\cite{hou2024relation} achieves similar improvements by incorporating location relationships across different decoder layers.
SAP-DETR~\cite{liu2023sap} proposes a modulated attention architecture augmented with additional salient points.
Furthermore, SAM-DETR~\cite{zhang2022accelerating}, HPR~\cite{zhao2024hybrid}, and MI-DETR~\cite{nan2025mi} integrate RoI-Align~\cite{ren2016faster} or multiple cross-attention mechanisms to enhance object queries in the transformer decoder.
Another line of research focuses on designing novel anchor box generation mechanisms~\cite{yao2021efficient,liu2022dab,meng2021conditional,wang2022anchor,hou2024salience,zhang2023dense}. For example, DAB-DETR~\cite{liu2022dab}, Conditional-DETR~\cite{meng2021conditional}, and Anchor-DETR~\cite{wang2022anchor} improve the transformer decoder's capability by associating anchor boxes with object queries.
Efficient-DETR~\cite{yao2021efficient}, Salience-DETR~\cite{hou2024salience}, and DDQ-DETR~\cite{zhang2023dense} further enhance detection by imposing a localization prior to generate stronger object proposals.
Additionally, lightweight DETR models~\cite{roh2021sparse,lin2022d,zhao2024detrs,li2023lite,chen2024lw} and localization-aware training objectives~\cite{zhang2023decoupled,liu2023detection,cai2023align,hou2024relation,pu2024rank} have been proposed to improve efficiency and accuracy.
In our work, we propose an instructive multi-route training mechanism that enhances the training of various DETR-like models.

\subsection{Training Detection Transformers}
Many typical object detectors~\cite{ren2016faster,tian2020fcos,zhang2020bridging,ross2017focal,ge2021ota,zheng2021yolox,feng2021tood} design various strategies to match each target with multiple predictions, thereby improving the detector's ability to learn robust representations~\cite{hdetr}.
In contrast, DETR-like detectors~\cite{detr,gao2022adamixer,chen2022recurrent,yang2022querydet,li2022exploring,cao2022cf,fang2024feataug} achieve end-to-end object detection through one-to-one matching, where each target is assigned to a single prediction, considering both localization and classification costs.
Nonetheless, DETR-like detectors with one-to-one matching frequently encounter slow convergence issues~\cite{hdetr,meng2021conditional,sun2021rethinking}.
To accelerate the training of vanilla DETR~\cite{detr}, several studies incorporate auxiliary training methods.
For instance, H-DETR~\cite{hdetr} leverages Hungarian Matching~\cite{kuhn1955hungarian} to construct one-to-many matching by duplicating targets.
DN-DETR~\cite{li2022dn} and DINO~\cite{dino} design multiple groups of denoising queries to accelerate the training of detection transformers.
Likewise, Group-DETR~\cite{chen2023group} utilizes a set of learnable queries as the auxiliary input for the transformer decoder.
DQ-DETR~\cite{cui2023learning} establishes the primary query group by dynamically fusing auxiliary queries.
StageInteractor~\cite{teng2023stageinteractor} introduces one-to-many matching by combining the label assignment results of different decoder layers.
SQR-DETR~\cite{chen2023enhanced} constructs auxiliary training by reusing object queries from previous decoder layers.
Co-DETR~\cite{zong2023detrs} adopts multiple query groups with different matching strategies to offer various supervision signals.
DAC-DETR~\cite{hu2024dac} develops a parallel decoder to learn a one-to-many assignment by eliminating self-attention.
MS-DETR~\cite{msdetr} proposes to supervise cross-attention output using one-to-many matching, while self-attention is supervised by one-to-one matching.
Additionally, several works~\cite{wang2024kd,chang2023detrdistill,huang2023teach} use knowledge distillation~\cite{hinton2015distilling,zhang2021delving} to enhance the training of DETR models.
Different from previous arts, we discuss the roles of each component in the transformer decoder in the multi-task framework, which is required to achieve one-to-one and one-to-many prediction.
We further propose an instructive multi-route training mechanism for the utilization of auxiliary training.

{\color{black}

\subsection{Mixture-of-Experts}
Mixture-of-Experts (MoE)~\cite{jacobs1991adaptive} aims to dynamically combine the knowledge of multiple experts while maintaining low computational cost during inference. This technique has been widely applied across various domains to scale up models, such as natural language processing~\cite{shazeer2017outrageously,fedus2022switch,liu2024deepseek} and computer vision~\cite{riquelme2021scaling,renggli2022learning,wu2022residual,zhang2023robust,fei2024scaling}. 
In the context of multi-task learning, M$_3$ViT~\cite{fan2022m3vit} and Mod-Squad~\cite{chen2023mod} introduce MoE-based models to effectively handle multiple tasks. AdaMV-MoE~\cite{chen2023adamv} enhances task specificity while maintaining efficiency through the use of partially shared experts. Furthermore, \cite{yang2024multi} proposes constructing experts with LoRA~\cite{hu2022lora} to address multiple dense prediction tasks.
}

\begin{figure*}[!thp]
    \centering
    \includegraphics[width=1.0\linewidth]{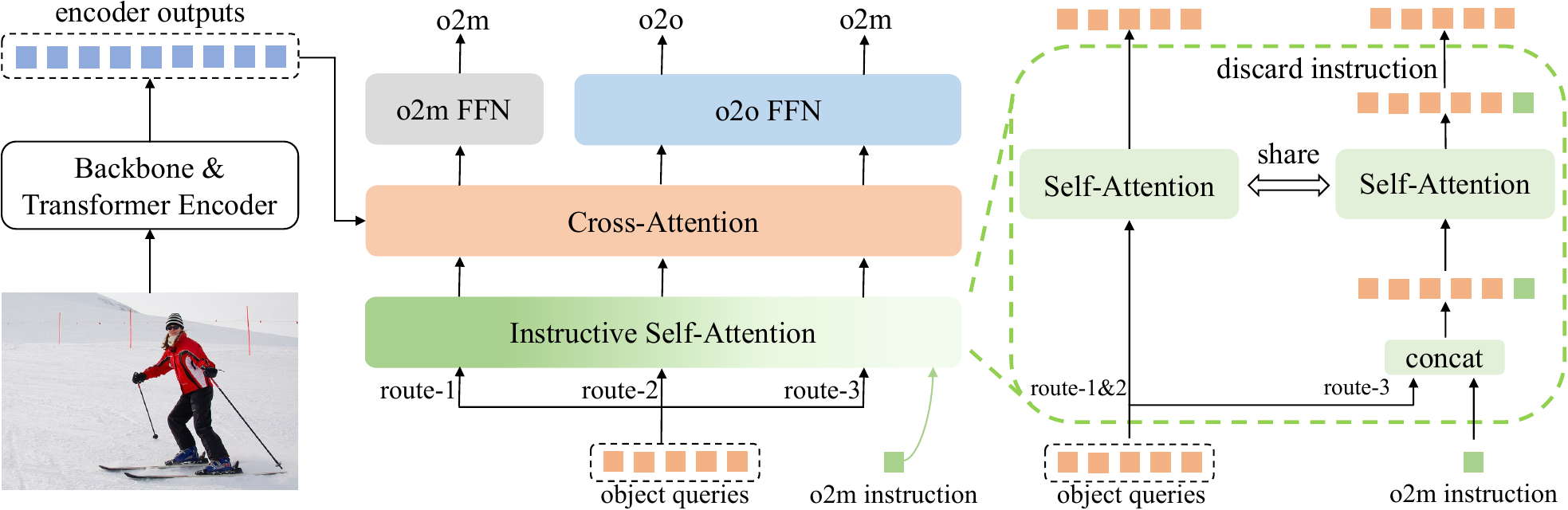}
    \caption{
    \textbf{Our proposed multi-route training method in Mr. DETR.} It includes three training routes: Route-1, Route-2, and Route-3. All three routes share the same object queries and detection heads for classification and regression. Route-2 serves as the primary route for one-to-one prediction, identical to the baseline models. Route-1 shares self-attention and cross-attention but uses an independent feed-forward network (o2m FFN) for one-to-many prediction. Route-3, sharing all components with the primary route, introduces a novel instructive self-attention, implemented by adding a learnable instruction token to the object queries to guide them and the subsequent network for one-to-many prediction. During inference, the auxiliary routes, Route-1 and Route-3, are discarded.
    }
    \label{fig:method}
    \vskip -0.2in
\end{figure*}

\section{Method}

\subsection{Preliminary}\label{sec:preliminary}
\noindent\textbf{DETR Architecture.}
DETR-like detectors typically consist of an image backbone, transformer encoder, and decoder.
The transformer encoder extracts features through self-attention among multi-scale image tokens.
A set of object queries $\mathbf{Q}=\{q_0, q_1, ..., q_n\}$ is fed into the transformer decoder, where classification and box regression heads derive the predicted classification $\mathbf{S}=\{s_0, s_1, ..., s_n\}$ and bounding boxes $\mathbf{B}=\{b_0, b_1, ..., b_n\}$ from the query output.
The decoder consists of $L$ stacked transformer layers, each containing self-attention, cross-attention, and a feed-forward network (FFN).
Self-attention is applied between object queries, cross-attention facilitates interaction between object queries and image features, and the FFN extracts features in the object queries.

\myPara{One-to-one Training Objective.}
Utilizing a one-to-one training objective~\cite{detr}, DETRs achieve end-to-end detection without the need for non-maximum-suppression (NMS).
Specifically, let $\mathbf{\Bar{B}}=\{\Bar{b}_0, \Bar{b}_1, ..., \Bar{b}_t\}$ and $\mathbf{\Bar{S}}=\{\Bar{s_0}, \Bar{s_1}, ..., \Bar{s_t}\}$ represent the ground-truth boxes and corresponding classes.
The matching cost between all possible prediction and ground-truth pairs is derived by considering both the classification cost and box costs.
Optimal matches are determined using bipartite matching~\cite{detr,kuhn1955hungarian}, denoted as $\sigma$.
The one-to-one training objective is expressed as:
\begin{equation}
    L = \sum_{i=0}^{t} L_{cls}(s_{\sigma(i)}, \Bar{s_i}) + L_{box}(b_{\sigma(i)}, \Bar{b_i}),\label{eq:loss}
\end{equation}
where the $L_{cls}$ and $L_{box}$ represent the classification and bounding box losses, respectively.

\myPara{One-to-many Training Objective.}
Traditional object detectors~\cite{ren2016faster,ross2017focal,tian2020fcos} typically use a one-to-many assignment strategy, assigning each ground truth box to multiple predictions based on specific criteria, followed by NMS to eliminate duplicated predictions.
In our work, we apply a straightforward one-to-many assignment strategy~\cite{ouyang2022nms} as in~\cite{msdetr,hu2024dac}, which considers localization quality and classification confidence, making it suitable for DETR-like detectors.
Specifically, the matching score $M_{ij}$ between the predictions $(s_i, b_i)$ and the ground truth $(\Bar{s_t}, \Bar{b_t})$ is defined as:
\begin{equation}
M_{ij} = \alpha \cdot s_i + (1 - \alpha) \cdot \texttt{IoU}(b_i, \Bar{b_j}),\label{eq:costmatrix}
\end{equation}
where $\texttt{IoU}$ computes the intersect-over-union between prediction box $b_i$ and the ground-truth box $\Bar{b_j}$.
Given a maximum number of positive candidates $K$ and an IoU threshold $\tau$, positive predictions can be determined.
First, up to $K$ predictions with the highest matching scores $M$ are selected as positive predictions.
Then, for each ground-truth box, predictions with an IoU lower than $\tau$ are filtered out.
The localization and classification losses are calculated as in~\eqref{eq:loss}.

\subsection{Multi-route Training}\label{sec:multiroute}
We aim to introduce the one-to-many assignment as an additional training strategy to enhance detection transformers.
First, we treat the detector with auxiliary one-to-many prediction as a multi-task framework, which simultaneously achieves both one-to-one and one-to-many predictions.
As discussed in~\secref{sec:introduction}, we empirically investigate the roles of each component in the transformer decoder within this framework and find:
\textbf{(i)} incorporating a one-to-many assignment significantly degrades the performance of the primary one-to-one prediction when all components are shared between two tasks (see~\tabref{tab:motivation}(2)).
We anticipate that \emph{it results from the interference between two tasks}.
For instance, a predicted box may be assigned as a positive prediction in the one-to-many assignment, while being assigned as a negative prediction in the one-to-one assignment.
\textbf{(ii)} any independent component in the decoder significantly benefits the primary one-to-one prediction route, even when other components are shared (see~\figref{fig:motivationfig} (b - d) \&~\tabref{tab:motivation} (3-5)).
This observation indicates that \emph{any independent component is capable of effectively mastering one-to-one and one-to-many training goals}, thereby resolving the conflict between these two tasks.
This is expected since shared components can extract common clues for two tasks, while independent components further distinguish different tasks;
\textbf{(iii)} an auxiliary training route with two independent components does not outperform the route with only one independent component (see~\tabref{tab:motivation}(6-8));
and \textbf{(iv)} combining the auxiliary route with independent self-attention and independent FFN achieves the highest performance among variants that combine different auxiliary training routes (see~\tabref{tab:motivation}(10)).

Building on these findings, our method includes three training routes, as shown in~\figref{fig:method}.
We share object queries, classification, and regression heads among the three routes.
Route-2 is the primary route for one-to-one prediction, which is identical to baseline models.
Route-1 and route-3 are auxiliary training routes used for one-to-many predictions, which are discarded during inference.
Thus, the auxiliary training routes in our method do not affect model architecture or inference time.

\myPara{Primary Route for One-to-one Prediction.} \
The architecture and training objective of route-2 in~\figref{fig:method} is the same as the baseline model.
Specifically, for Route-2, given object queries $\mathbf{Q}=\{q_0, q_1, ..., q_{n-1}\}$, the query output is defined as:
\begin{equation}
    \hat{\mathbf{Q}_{2}} = (\texttt{FFN}_{o2o} \circ \texttt{CA} \circ \texttt{SA})(\mathbf{Q}),
\end{equation}
where $\texttt{SA}$, $\texttt{CA}$ and $\texttt{FFN}_{o2o}$ represent self-attention, cross-attention and FFN, respectively.
The query output of Route-2 is supervised by one-to-one assignment in~\eqref{eq:loss}.
During inference, Route-2 is retained to achieve one-to-one prediction without any additional cost.

\myPara{Auxiliary Route with Independent FFN.} \
As depicted in~\figref{fig:method}, we integrate an auxiliary training route, referred to as Route-1, featuring a separate FFN into our approach.
Specifically, we directly employ an independent FFN, $\texttt{FFN}_{o2m}$ in Route-1 sharing all self-attention and cross-attention components with the primary route.
Due to the straightforward architecture and efficient parameter utilization of the FFN, we maintain its independence without further modification.
The query output $\hat{\mathbf{Q}_{1}}$ of Route-1 can be written as:
\begin{equation}
    \hat{\mathbf{Q}_{1}} = (\texttt{FFN}_{o2m} \circ \texttt{CA} \circ \texttt{SA})(\mathbf{Q}).
\end{equation}
This output is supervised by a one-to-many assignment.

\myPara{Auxiliary Route with Instructive Self-Attention.}
In~\secref{sec:introduction}, we discuss the motivation for incorporating independent self-attention in Route-3 as shown in~\figref{fig:motivationfig}(e), designed for one-to-many prediction.
To further reduce trainable parameters and enhance parameter sharing with the primary route, we propose an innovative instruction mechanism as shown in~\figref{fig:method}.
This mechanism guides shared object queries to enable one-to-many prediction.
The query output $\hat{\mathbf{Q}_{3}}$ of Route-3 is written as:
\begin{equation}
    \hat{\mathbf{Q}_{3}} = (\texttt{FFN}_{o2o} \circ \texttt{CA} \circ \texttt{InstructSA}) (\mathbf{Q}),\label{eq:route-3}
\end{equation}
where $\texttt{InstrcutSA}$ denotes our proposed instructive self-attention which shares parameters with the self-attention in the other two routes.
The output $\hat{\mathbf{Q}_{3}}$ is supervised by one-to-many assignment.
Next, we will present the details of our instructive self-attention $\texttt{InstrcutSA}$ design.

\subsection{Instructive Self-Attention}\label{sec:insattnsec}
\begin{figure}
    \centering
    \includegraphics[width=\linewidth]{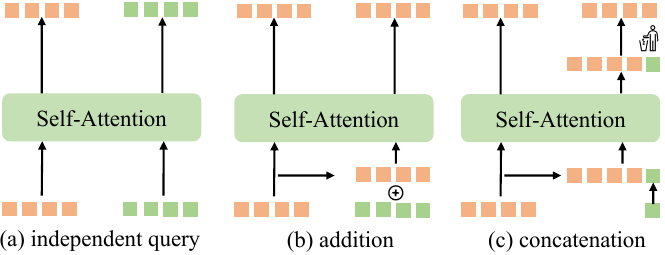}
    \caption{\textbf{Various implementations of instructive self-attention.}}\label{fig:designsofattn}
    \vskip -0.2in
\end{figure}

As described in~\secref{sec:multiroute}, Route-3 is realized using an instructive self-attention, which incorporates learnable instruction tokens $\mathbf{Q^{ins}}$ into object queries $\mathbf{Q}$, creating a combined sequence $\mathbf{\hat{\mathbf{Q}}^{ins}}$.
Self-Attention is performed on this combined sequence.
In this section, we examine distinct implementations of instructive self-attention aimed at informing object queries to realize one-to-many predictions, sharing self-attention parameters with the primary route. 
The simplest approach, depicted in~\figref{fig:designsofattn}(a), involves using a separate set of queries as inputs to facilitate one-to-many predictions.
To improve the compatibility of object queries across different routes, we introduce learnable tokens that serve as instruction, delineating the prediction objectives. 
As depicted in~\figref{fig:designsofattn}(b), these instruction tokens are incorporated into shared object queries through addition.
This approach necessitates a fixed number of instruction tokens equivalent to the query count. 
Unlike the addition approach, our method adapts instruction tokens through concatenation, thereby providing greater flexibility. 
As illustrated in~\figref{fig:designsofattn}(c), this flexibility extends not just to the number of instruction tokens, but also allows these learned tokens to dynamically transmit information to object queries via self-attention.
Self-attention is performed on the combined sequence $\mathbf{\hat{\mathbf{Q}}^{ins}}$, but the corresponding outputs of the instructive tokens are discarded after self-attention because they are not intended for object localization.
Further details can be found in~\eqref{eq:inssa}.
In this way, the instruction tokens provide effective and adaptable guidance to Route-3, enabling it to make one-to-many predictions while utilizing shared parameters with the primary route.
In \secref{sec:ablationstudy}, we evaluate the performance of various implementations.

Specifically, we build $m$ learnable tokens $\mathbf{Q^{ins}}=\{q_0^{ins}, q_1^{ins}, ..., q_{m-1}^{ins}\}$, called instruction tokens.
Initially, these instruction tokens are attached to the input sequence of self-attention by concatenation, forming a composite set of input queries
$\mathbf{\hat{Q}^{ins}} = \{q_0^{ins}, q_1^{ins}, ..., q_{m-1}^{ins}, q_0, q_1, ..., q_{n-1}\}$, resulting in a length of $m+n$.
Subsequently, self-attention is performed on these combined queries.
As the attached instruction tokens are not utilized for object localization, their outputs are discarded post self-attention, serving solely to convey information to object queries.
The output $\hat{\mathbf{Q}}_3$ of Route-3 is written as:
\begin{equation}
    \hat{\mathbf{Q}_{3}} = (\texttt{FFN}_{o2o} \circ \texttt{CA} \circ \texttt{R} \circ \texttt{SA}) (\mathbf{\hat{\mathbf{Q}}^{ins}}),\label{eq:inssa}
\end{equation}
where the function $\texttt{R}$ eliminates the outputs of instruction tokens after self-attention.
Notably, all components of Route-3, including self-attention, cross-attention, and FFN, share parameters with the primary route.
Instruction tokens effectively guide object queries and subsequent modules to achieve one-to-many prediction.
The shared parameters between the auxiliary and primary routes benefit the one-to-one prediction of the primary route.

\myPara{Discussion.} \
Large language models have demonstrated impressive capabilities~\cite{brown2020language} when equipped with meticulously crafted prompts~\cite{shin2020autoprompt,jiang2020can}.
Subsequent researches treat these prompts as learnable tokens, achieving success in both natural language processing~\cite{li2021prefix,lester2021power,liu2021p} and computer vision applications~\cite{zhou2022learning,ju2022prompting,wang2022learning,li2025imove}.
While prompt tuning~\cite{jia2022visual,shen2024multitask,yoo2023improving} and our proposed instructive self-attention both involve attaching learnable tokens to the input sequence, our instruction token uniquely facilitates distinguishing between one-to-one and one-to-many assignments, rather than merely fine-tuning the model. We name our method Mr. DETR, due to its \underline{M}ulti-\underline{r}oute training nature for \underline{DETR}-based detectors.

\begin{figure*}[!thp]
    \centering
    \includegraphics[width=1.0\linewidth]{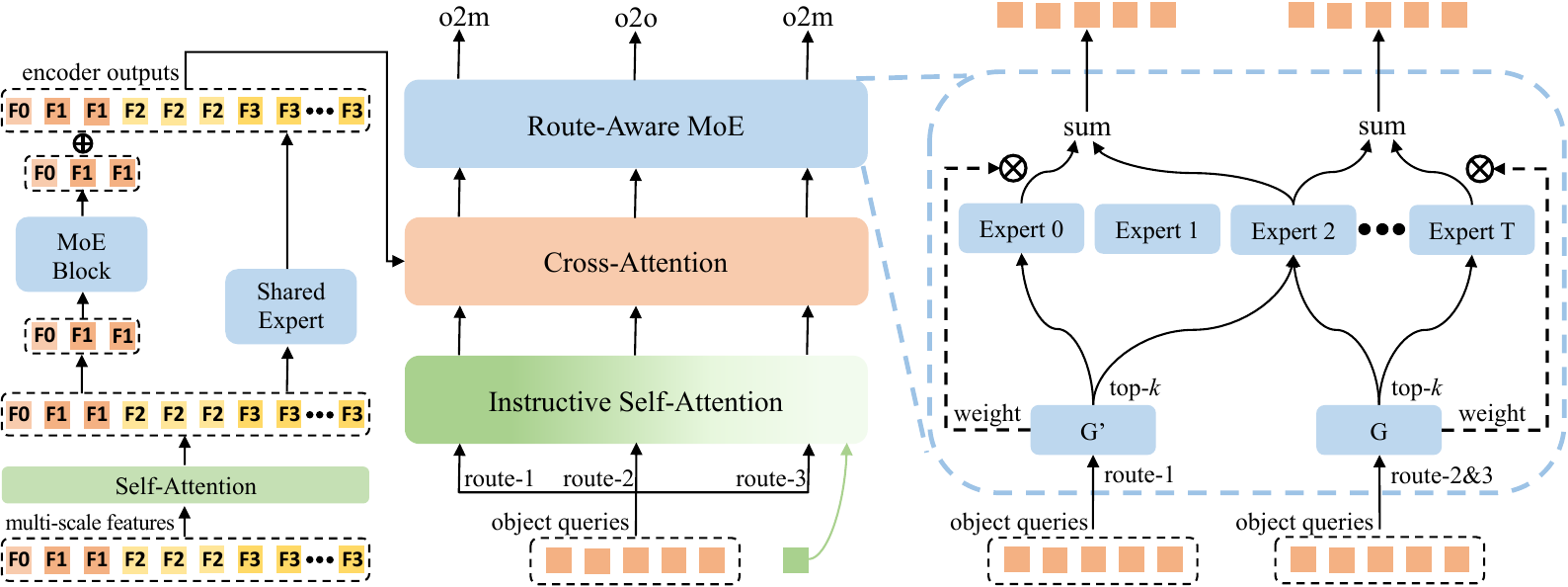}
    \caption{\color{black}{\textbf{Illustration of the proposed Mr. DETR++ architecture.} 
Our approach replaces the two independent FFNs in Mr. DETR with a route-aware MoE, enabling knowledge sharing and task-specific specialization across routes. Two distinct gating functions, $\texttt{G}$ and $\texttt{G'}$, are employed: $\texttt{G}$ governs Route-1, while $\texttt{G'}$ is shared between Route-2 and Route-3. The MoE block is further integrated into the transformer encoder. To reduce computational cost, the MoE block is applied to low-scale features, with all tokens sharing a single expert.}}
    \label{fig:mrdetrppmethod}
    \vskip -0.2in
\end{figure*}

{\color{black}
\subsection{From Mr. DETR to Mr. DETR++}
As illustrated in~\figref{fig:motivationfig}(g), to enable knowledge sharing between the two independent FFNs used in Mr. DETR, we propose a route-aware MoE to replace the FFNs.
We further extend the MoE architecture to the transformer encoder to enhance feature representation. 
To mitigate computational overhead, we introduce a scale-aware MoE applied to low-scale features.
Additionally, we incorporate a localization-aware score to improve classification calibration.
This extended method is called Mr. DETR++.

\myPara{Route-aware Mixture-of-Experts in Decoder.}
The two independent FFNs used in Mr. DETR can be regarded as distinct experts tailored to separate tasks, representing a special case of the MoE paradigm.
Thus, as shown in~\figref{fig:mrdetrppmethod}, we propose a route-aware MoE to replace the FFNs, enabling knowledge sharing and task-specific specialization across routes.
Specifically, three routes share a set of $t$ experts $E=\{e_0,e_1,\cdot\cdot\cdot,e_{t-1}\}$ within the MoE framework.
To prevent conflicts between routes, Route-2 and Route-3 share a gating function $\texttt{G}(\cdot)$, while Route-1 employs an independent gating function $\texttt{G'}(\cdot)$.

Formally, the query output $\hat{\mathbf{Q}_{1}}$ for Route-1 in Mr. DETR++ is expressed as:
\begin{equation}
    \hat{\mathbf{Q}_{1}} = (\texttt{E} \circ \texttt{G'} \circ \texttt{CA} \circ \texttt{SA})(\mathbf{Q}),
\end{equation}
where \texttt{SA} denotes self-attention, \texttt{CA} denotes cross-attention, $\texttt{G'}$ is the gating function, and $\texttt{E}$ represents the expert computation.
Similarly, the query output $\hat{\mathbf{Q}_{2}}$ for the one-to-one prediction in Route-2 is given by:
\begin{equation}
    \hat{\mathbf{Q}_{2}} = (\texttt{E} \circ \texttt{G} \circ \texttt{CA} \circ \texttt{SA})(\mathbf{Q}).
\end{equation}
For Route-3, equipped with instructive self-attention, the query output $\hat{\mathbf{Q}_{3}}$ is:
\begin{equation}
    \hat{\mathbf{Q}_{3}} = (\texttt{E} \circ \texttt{G} \circ \texttt{CA} \circ \texttt{InstructSA})(\mathbf{Q}).
\end{equation}
This route-aware MoE design facilitates knowledge sharing while mitigating potential conflicts.

For the MoE block, we use sparse activation of the top-$k$ experts.
The gating function $\texttt{G}(\cdot)$ (or $\texttt{G'}(\cdot)$ for Route-1) computes scores for each expert, and only the top-$k$ experts with the highest scores are activated for processing the query.
For a given query $\mathbf{Q}$ and gating function $\texttt{G}(\cdot)$, the gating output is a vector of scores $\mathbf{c} = \texttt{G}(\mathbf{Q})$, where $c_i$ represents the score for expert $e_i$. The top-$k$ expert indices are selected as:
\begin{equation}
    \mathcal{T}_k = \text{TopK}(\mathbf{s}, k),
\end{equation}
where $\text{TopK}(\mathbf{s}, k)$ returns the indices of the $k$ experts with the highest scores.
The output of the MoE block for the query $\mathbf{Q}$ is then computed as a weighted sum of the selected experts' outputs:
\begin{equation}
    \hat{\mathbf{Q}} = \sum_{i \in \mathcal{T}_k} c_i \cdot e_i(\mathbf{Q}),\label{eq:moe}
\end{equation}
where $e_i(\mathbf{Q})$ is the output of expert $e_i$ for the input query $\mathbf{Q}$, and $c_i$ is the corresponding gating score.

\myPara{Scale-aware Mixture-of-Experts in Encoder.}
We further extend the MoE framework to the transformer encoder to enhance feature representations.
Typically, the encoder processes a sequence of flattened feature tokens from multiple scales, $F = \{f_0, f_1, \dots, f_{d-1}\}$, where $f_i$ represents image tokens at scale $i$.
Directly applying MoE to this long sequence would significantly increase computational cost.
To address this, we propose a scale-aware MoE that applies a shared expert to all tokens and a specialized MoE block to low-scale features.

Assuming the token sequence length increases with scale index, similar to~\eqref{eq:moe}, we apply the MoE to the first $\eta$ scales (low-scale features), and a shared FFN is applied to all tokens in $F$.
The final features are computed by combining outputs for the first $\eta$ scales and retaining the FFN output for higher scales:
\begin{equation}
    \hat{f}_j' = \begin{cases} 
        \text{FFN}(f_j) + \text{MoE}(f_j), & j = 0, 1, \dots, \eta-1, \\
        \text{FFN}(f_j), & j = \eta, \dots, d-1.
    \end{cases}
\end{equation}
This scale-aware MoE enhances representation capacity for low-scale features while maintaining computational efficiency for higher scales.

\myPara{Localization-aware Score Calibration.}
Relying solely on classification scores can lead to low-quality bounding boxes being ranked higher than high-quality ones, resulting in suboptimal performance.
This issue has been widely studied~\cite{zhang2021varifocalnet,liu2023detection,pu2024rank,ye2023cascade,cai2023align}.
Unlike Cascade-DETR~\cite{ye2023cascade}, which introduces a class-agnostic IoU score branch alongside the classification score, we propose a class-aware IoU score, inspired by VarifocalNet~\cite{zhang2021varifocalnet}, to calibrate the classification score.
During training, in contrast to Stable-DINO~\cite{liu2023detection} and Align-DETR~\cite{cai2023align}, the label assignment procedure excludes the IoU score, simplifying the prediction target and label assignment.
We employ the VFL Loss~\cite{zhang2021varifocalnet} to learn the class-aware IoU score, where the target score is empirically lifted by $s_{target} = iou^{0.75}$.
At inference, we compute a calibrated score by combining the classification and IoU scores:
\begin{equation}
    s_{\text{calib}} = s_{\text{cls}}^{\phi} \cdot s_{\text{iou}}^{1-\phi}, \label{eq:calib_score}
\end{equation}
where $s_{\text{cls}} \in [0, 1]$ is the classification score, $s_{\text{pred}} \in [0, 1]$ is the predicted IoU score, and $\phi$ is to balance the contributions of classification and localization confidence.
This calibration is easy to use and does not affect the original model’s training process.
}

\newcommand{\major}[1]{\textcolor{blue}{#1}}
\begin{table*}[!thp]
\centering
\setlength{\tabcolsep}{6pt} 
\caption{\textbf{The performance on the COCO 2017~\cite{coco} validation set.} All models are based on the ResNet-50~\cite{he2016deep} backbone.}\label{tab:mainresults}
\resizebox{0.9\textwidth}{!}{
\begin{tabular}{l|c|c|cc|cccccc}
\toprule
Model & w/ Mr. DETR & w/ Mr. DETR++ & Epochs & Queries & AP & AP$_{50}$ & AP$_{75}$ & AP$_s$ & AP$_m$ & AP$_l$ \\

\midrule
Deformable DETR++~\cite{deformable} &  && 12 & 300 & 47.0 & 65.3 & 51.0 & 30.1 & 50.5 & 60.7 \\
H-DETR~\cite{hdetr} &  && 12 & 300 & 48.7 & 66.4 & 52.9 & 31.2 & 51.5 & 63.5 \\
MS-DETR~\cite{msdetr} &  && 12 & 300 & 48.8 & 66.2 & 53.2 & 31.5 & 52.3 & 63.7 \\
\rowcolor{blue!10}Deformable-DETR++~\cite{deformable} & \ding{52} && 12 & 300 & 49.5 (\textcolor{blue}{+2.5}) & 67.0 & 53.7 & 32.1 & 52.5 & 64.7 \\ 
\rowcolor{blue!10}H-DETR~\cite{hdetr} & \ding{52} &  & 12 & 300 & 49.8 (\textcolor{blue}{+1.1}) & 67.3 & 54.5 & 31.6 & 53.0 & 65.6 \\ 
\rowcolor{blue!10}Deformable-DETR++~\cite{deformable} &  & \ding{52} & 12 & 300 & \textbf{51.0} (\textcolor{blue}{+4.0}) & \textbf{67.8} & \textbf{55.6} & \textbf{33.2} & \textbf{54.8} &\textbf{66.6} \\

\midrule
Deformable-DETR++~\cite{deformable} &  && 12 & 900 & 47.6 & 65.8 & 51.8 & 31.2 & 50.6 & 62.6 \\ 
DINO~\cite{dino} &  && 12 & 900 & 49.0 & 66.6 & 53.5 & 32.0 & 52.3 & 63.0 \\
MS-DETR~\cite{msdetr} &  && 12 & 900 & 50.0 & 67.3 & 54.4 & 31.6 & 53.2 & 64.0 \\
DAC-DETR~\cite{hu2024dac} &  && 12 & 900 & 49.3 & 66.5 & 53.8 & 31.4 & 52.4 & 64.1 \\
\rowcolor{blue!10}Deformable-DETR++~\cite{deformable} & \ding{52}  && 12 & 900 & 50.7 (\textcolor{blue}{+3.1}) & 68.2 & 55.4 & 33.6 & 54.3 & 64.6 \\
\rowcolor{blue!10}Deformable-DETR++~\cite{deformable} & & \ding{52} & 12 & 900 & \textbf{51.8} (\textcolor{blue}{+4.2}) & \textbf{68.9} & \textbf{56.7} & \textbf{35.6} & \textbf{55.9} & \textbf{67.0} \\

\midrule
Deformable-DETR++~\cite{deformable} &  && 24 & 900 & 49.8 & 67.0 & 54.2 & 31.4 & 52.8 
& 64.1 \\
DINO-DETR~\cite{dino} & && 24 & 900 & 50.4 & 68.3 & 54.8 &  33.3  & 53.7 & 64.8 \\
MS-DETR~\cite{msdetr} &  && 24 & 900 & 50.9 & 68.4 & 56.1 &  34.7 & 54.3 & 65.1\\
DAC-DETR~\cite{hu2024dac} &  && 24 & 900 & 50.5 & 67.9 & 55.2 & 33.2 & 53.5 & 64.8 \\
\rowcolor{blue!10}Deformable-DETR++~\cite{deformable} & \ding{52}  && 24 & 900 & 51.4 (\textcolor{blue}{+1.6}) & 69.0 & 56.2 & 34.9 & 54.8 & 66.0 \\
\rowcolor{blue!10}Deformable-DETR++~\cite{deformable} & & \ding{52} & 24 & 900 & \textbf{52.0} (\textcolor{blue}{+2.2}) & \textbf{69.1} & \textbf{56.4} & \textbf{35.4} & \textbf{55.5} & \textbf{66.6} \\

\midrule
DINO~\cite{dino} &  && 12 & 900 & 49.0 & 66.6 & 53.5 & 32.0 & 52.3 & 63.0 \\
Salience-DETR~\cite{hou2024salience} &  && 12 & 900 & 49.2 & 67.1 & 53.8 & 32.7 & 53.0 & 63.1 \\
Group-DETR~\cite{chen2023group} &  && 12 & 900 & 49.8 & - & - & 32.4 & 53.0 & 64.2 \\
MS-DETR~\cite{msdetr} &  && 12 & 900 & 50.3 & 67.4 & 55.1 & 32.7 & 54.0 & 64.6 \\
DAC-DETR~\cite{hu2024dac} & & & 12 & 900 & 50.0 & 67.6 & 54.7 & 32.9 & 53.1 & 64.2 \\
Cascade-DETR~\cite{ye2023cascade} &  && 12 & 900 & 49.7 & 67.1 & 54.1 & 32.4 & 53.5 & 65.1 \\
Align-DETR~\cite{cai2023align} &  && 12 & 900 & 50.2 & 67.8 & 54.4 & 32.9 & 53.3 & 65.0 \\
Rank-DETR~\cite{pu2024rank} & & & 12 & 900 & 50.4 & 67.9 & 55.2 & 33.6 & 53.8 & 64.2 \\
Stable-DINO~\cite{liu2023detection} & & & 12 & 900 & 50.4 & 67.4 & 55.0 & 32.9 & 54.0 & 65.5 \\
EASE-DETR~\cite{gao2024ease} &  && 12 & 900 & 49.7 &  67.5 & 54.3 & 32.7 & 52.9 &  64.1 \\
\rowcolor{blue!10}DINO~\cite{dino} & \ding{52} & & 12 & 900 & 50.9 (\textcolor{blue}{+1.9}) & 68.4 & 55.6 & 34.6 & 53.8 & 65.2 \\
\rowcolor{blue!10}DINO~\cite{dino} & &\ding{52} & 12 & 900 & \textbf{52.2} (\textcolor{blue}{+3.2}) & \textbf{69.7} & 56.3 & \textbf{35.9} & \textbf{55.8} & \textbf{67.2} \\
\rowcolor{blue!10}Align-DETR~\cite{cai2023align} & \ding{52} & &12 & 900 & 51.4 (\textcolor{blue}{+1.2}) & 68.6 & 55.7 & 33.8 & 54.7 & 66.3 \\
\rowcolor{blue!10}Align-DETR~\cite{cai2023align} && \ding{52} &12 & 900 & \textbf{52.2} (\textcolor{blue}{+2.0}) & \textbf{69.7} & \textbf{56.8} & 34.8 & \textbf{55.8} & 66.7 \\

\midrule
DINO~\cite{dino} & & & 24 & 900 & 50.4 & 68.3 & 54.8 & 33.3 & 53.7 & 64.8 \\
Align-DETR~\cite{cai2023align} & & & 24 & 900 & 51.3 & 68.2 & 56.1 & 35.5 & 55.1 & 65.6 \\
MS-DETR~\cite{msdetr} && & 24 & 900 & 51.7 & 68.7 & 56.5 & 34.0 & 55.4 & 65.5 \\
DDQ-DETR~\cite{zhang2023dense} && & 24 & 900 & 52.0 & 69.5 & \textbf{57.2} & 35.2 & 54.9 & 65.9 \\
Stable-DINO~\cite{liu2023detection} &  & & 24 & 900 & 51.5 & 68.5 & 56.3 & 35.2 & 54.7 & 66.5 \\
DAC-DETR~\cite{hu2024dac} &  & & 24 & 900 & 51.2 & 68.9 & 56.0 & 34.0 & 54.6 & 65.4 \\
\rowcolor{blue!10}DINO~\cite{dino} & \ding{52}   & &24 & 900 & 51.7 (\textcolor{blue}{+1.3}) & 69.2 & 56.4 & 34.1 & 55.1 & 65.8\\
\rowcolor{blue!10}DINO~\cite{dino} & &\ding{52}  &24 & 900 & 52.3 (\textcolor{blue}{+1.9}) & 69.9 & 56.7 & \textbf{36.3} & \textbf{56.1} & 66.7 \\
\rowcolor{blue!10}Align-DETR~\cite{cai2023align} & \ding{52}  & & 24 & 900 &  52.3 (\textcolor{blue}{+1.0}) & 69.5 & 56.7 & 35.2 & 56.0 & \textbf{67.0}\\
\rowcolor{blue!10}Align-DETR~\cite{cai2023align} & &\ding{52} & 24 & 900 & \textbf{52.4} (\textcolor{blue}{+1.1}) & \textbf{70.1} & \textbf{56.9} & \textbf{36.3} &  55.9 & 66.9 \\

\bottomrule

\end{tabular}
}

\vskip -0.2in
\end{table*}

\section{Experiments}
\subsection{Setup}

{
\color{black}
\subsubsection{Datasets and Evaluation}
We conduct extensive experiments on object detection, instance segmentation, and panoptic segmentation using three popular datasets (see~\appref{appdix:datasets}): COCO 2017~\cite{coco}, NuScenes~\cite {caesar2020nuscenes}, and Objects365~\cite{shao2019objects365}.
In line with existing research~\cite{deformable,dino,hdetr}, we conduct evaluations on the COCO 2017 validation set, providing results in terms of standardized metrics, namely average precision (mAP, AP$_{50}$, AP$_{75}$) at various IoU thresholds. 
AP$_s$, AP$_m$, and AP$_l$ refer to the average precision for small, medium, and large objects, respectively. 
For all experiments with 300 and 900 queries, we evaluate the performance according to top-100 and top-300 predictions, respectively.
For the instance segmentation task, we report the mask AP and box AP, respectively.
Panoptic segmentation is evaluated using the Panoptic Quality (PQ) metric~\cite{kirillov2019panoptic}.
We report the PQ for `thing' and `stuff' classes, respectively.
Besides, we evaluate the box AP and mask AP for `thing' classes.
}

\subsubsection{Implement Details}
We conduct all experiments based on the ResNet-50~\cite{he2016deep} and Swin-L~\cite{liu2021swin} as the backbone pretrained on the ImageNet~\cite{deng2009imagenet}.
In all experiments, we apply the same data augmentations as~\cite{deformable,dino,hdetr,hu2024dac,msdetr}.
AdamW~\cite{loshchilov2017decoupled} is employed as the optimizer, with the initial learning rate and weight decay established at 2e-4 and 1e-4, respectively.
We utilize a batch size of 16 for training all the models.
During training schedules of 12 and 24 epochs, the learning rate is reduced by a factor of 0.1 following the 11th and 20th epochs, respectively.
In all experiments with instructive self-attention, we set the number of instruction tokens to 10 (see~\appref{appendix:inssa}).
We configure Mr. DETR++ with four experts, activating the top two experts for both the encoder and decoder (see~\appref{appendix:moedesign}).
Within the one-to-many assigner, the hyper-parameters are configured as follows: $K = 6$, $\alpha = 0.3$, and $\tau = 0.4$ (see~\appref{appendix:assignmentshyper}).
For models based on DINO~\cite{dino}, we use 100 contrastive denoising queries.

\subsection{Object Detection}
\subsubsection{MS COCO 2017}
{\color{black}
We conduct extensive experiments on the COCO 2017 dataset.
In~\tabref{tab:mainresults}, we present the performance of our method across various baselines, including Deformable-DETR++~\cite{deformable} with either 300 or 900 queries, DINO~\cite{dino} and Align-DETR~\cite{cai2023align}.
Specifically, using Deformable-DETR++~\cite{deformable} with 300 queries, Mr. DETR achieves a 49.5\% mAP, providing a 2.5\% improvement without additional inference cost.
It surpasses other variants, such as H-DETR~\cite{hdetr}, MS-DETR~\cite{msdetr} and DAC-DETR~\cite{hu2024dac}, by approximately 0.7\% in mAP.
Mr. DETR++ reaches 51.0\% mAP, outperforming the baseline model by 4\%.
Based on a strong baseline model like Deformable-DETR++~\cite{deformable} with 900 queries, our Mr. DETR and Mr. DETR++ reach a 50.7\% mAP and 51.8\% under a 12-epoch training schedule, respectively.
Extending to 24 epochs, Mr. DETR and Mr. DETR++ achieve a 51.4\% and 52.0\% mAP, outperforming the baseline by 1.6\% and 2.2\%, respectively.
Compared to other variants with one-to-many auxiliary training, Mr. DETR++ exceeds DAC-DETR~\cite{hu2024dac} and MS-DETR~\cite{msdetr} by 1.5\% and 1.1\%, respectively.

\begin{table}[!thp]
\centering
\setlength{\tabcolsep}{5pt} 
\small
\caption{\textbf{The performance on the COCO 2017~\cite{coco} validation set.} Our model is based on DINO~\cite{dino} with IA-BCE loss~\cite{cai2023align}. All models are trained based on Swin-L~\cite{liu2021swin} backbone for 12 epochs with 900 queries, except for H-DETR~\cite{hdetr}, which uses 300 queries.}\label{tab:swinres}
\resizebox{0.49\textwidth}{!}{
\begin{tabular}{l|cccccc}
\toprule
Model & AP & AP$_{50}$ & AP$_{75}$ & AP$_s$ & AP$_m$ & AP$_l$ \\
\midrule
H-DETR~\cite{hdetr}  & 55.9 & 75.2 & 61.0 & 39.1 & 59.9 & 72.2 \\
DINO~\cite{dino}  & 56.8 & 75.4 & 62.3 & 41.1 & 60.6 & 73.5 \\
Co-DETR~\cite{zong2023detrs} & 56.9 & 75.5 & 62.6 & 40.1 & 61.2 & 73.3 \\
Salience-DETR~\cite{hou2024salience} & 56.5 & 75.0 & 61.5 & 40.2 & 61.2 & 72.8 \\
Rank-DETR~\cite{pu2024rank}  & 57.6 & 76.0 &  63.4  & 41.6 & 61.4 & 73.8 \\ 
DAC-DETR~\cite{hu2024dac} & 57.3 & 75.7 & 62.7 & 40.1 & 61.5 & 74.4 \\
EASE-DETR~\cite{gao2024ease} &  57.8 & \textbf{76.7} & 63.3 & 40.7 & 61.9 & 73.7 \\
Stable-DINO~\cite{liu2023detection}  & 57.7 & 75.7 & 63.4 & 39.8 & 62.0 & 74.7 \\
Relation-DETR~\cite{hou2024relation} & 57.8&  76.1 & 62.9 & 41.2&  62.1&  74.4 \\
\rowcolor{blue!10}Mr. DETR (\textbf{ours}) & 58.4 & 76.3 & 63.9 & 40.8 & 62.8 & \textbf{75.3} \\
\rowcolor{blue!10}Mr. DETR++ (\textbf{ours}) & \textbf{58.7} & 76.5 & \textbf{64.0} & \textbf{42.2} & \textbf{62.9} & 75.2\\

\bottomrule

\end{tabular}
}
\vskip -0.1in
\end{table}

Based on DINO~\cite{dino}, our proposed Mr. DETR and Mr. DETR++ achieve 50.9\% and 52.2\% AP, respectively, over 12 epochs, surpassing variants such as Group-DETR~\cite{chen2023group}, DAC-DETR~\cite{hu2024dac} and MS-DETR~\cite{msdetr}.
When equipped with Mr. DETR and Mr. DETR++, Align-DETR yields AP improvements of 1.2\% and 2.0\%, respectively.
Our experiments consistently demonstrate performance enhancements across different baseline models.
Additional experiments utilizing the Swin-L~\cite{liu2021swin} backbone, as shown in~\tabref{tab:swinres}, show that Mr. DETR and Mr. DETR++ attain 58.4\% and 58.7\% AP, respectively, outperforming other variants.
These experimental results highlight the efficacy of our proposed method.
}

\begin{table}[!t]
    \caption{\textbf{Experiments on the large-scale Obejcts365 dataset}~\cite{shao2019objects365}. Deformable-DETR++ (baseline) and our model use the ResNet-50~\cite{he2016deep} backbone. All models are trained for 12 epochs with 900 queries.}\label{tab:objects365}
    \centering
    \small
    \setlength{\tabcolsep}{4pt}
    \begin{tabular}{l|cccccc}
    \toprule
      Models  & AP & AP$_{50}$ & AP$_{75}$ & AP$_{s}$ & AP$_{m}$ & AP$_{l}$ \\ \midrule
      Baseline~\cite{deformable} & 30.4 & 40.8 & 33.1 & 16.1 & 30.1 & 39.1 \\  
      \rowcolor{blue!10} w/ Mr. DETR & 32.7 (\textcolor{blue}{+2.3}) & 42.7 & 35.8  & 17.1  & 32.3  & 42.6  \\ 
      \rowcolor{blue!10} w/ Mr. DETR++ & \textbf{34.9} (\textcolor{blue}{+4.5}) & 45.4 & \textbf{38.2} & \textbf{18.3}  & \textbf{34.3} & \textbf{45.7}  \\ 
      \bottomrule
    \end{tabular}
    \vskip -0.2in
\end{table}

\begin{table}[t]
    \caption{\textbf{Experiments on the NuImages dataset~\cite{caesar2020nuscenes}.} Deformable-DETR++ (baseline) and our models use the ResNet-50~\cite{he2016deep} backbone. All models are trained for 12 epochs with 900 queries.}\label{tab:nuimages}
    \centering
    \small
    \setlength{\tabcolsep}{4pt}
    \begin{tabular}{l|cccccc}
    \toprule
      Models & AP & AP$_{50}$ & AP$_{75}$ & AP$_{s}$ & AP$_{m}$ & AP$_{l}$ \\ \midrule
      Baseline~\cite{deformable} & 48.5 & 75.1 & 52.3 & 27.4 & 46.1  & 63.0 \\  
      \rowcolor{blue!10} w/ Mr. DETR  & 51.2 (\textcolor{blue}{+2.7}) & \textbf{76.4} & 55.8 & \textbf{28.3} & 48.1 & 66.4  \\ 
      \rowcolor{blue!10} w/ Mr. DETR++ & \textbf{52.3} (\textcolor{blue}{+3.8}) & 75.8 & \textbf{56.5} & 25.5 & \textbf{49.0} & \textbf{69.0} \\ 
      \bottomrule
    \end{tabular}
\end{table}

\begin{table}[!t]
\caption{\textbf{Instance segmentation results on the COCO 2017~\cite{coco} validation set.} All experiments are based on Deformable-DETR++~\cite{deformable} with 300 queries and ResNet-50~\cite{he2016deep} as backbone.}\label{tab:insseg}
\centering
\setlength{\tabcolsep}{4pt} 
\small
\begin{tabular}{l|ccc}
\toprule
Models & Epochs & Mask AP & Box AP \\
\midrule
Deformable-DETR++~\cite{deformable} & 12 & 32.4 & 46.5 \\
\rowcolor{blue!10}w/ Mr. DETR & 12 & 36.0 (\textcolor{blue}{+3.6}) & 49.5 (\textcolor{blue}{+3.0}) \\
\rowcolor{blue!10}w/ Mr. DETR++ & 12 & \textbf{37.7} (\textcolor{blue}{+5.3}) & \textbf{50.8} (\textcolor{blue}{+4.3}) \\
\midrule
Deformable-DETR++~\cite{deformable} & 24 & 35.1 & 48.6 \\
\rowcolor{blue!10}w/ Mr. DETR & 24 & 37.6 (\textcolor{blue}{+2.5}) & 50.3 (\textcolor{blue}{+1.7}) \\
\rowcolor{blue!10}w/ Mr. DETR++ & 24 & \textbf{39.0} (\textcolor{blue}{+3.9}) & \textbf{51.1} (\textcolor{blue}{+2.5}) \\

\bottomrule
\end{tabular}
\vskip 0.0in
\end{table}

{\color{black}
\subsubsection{Objects365}
To further verify the scalability of our method on the large-scale dataset, we conduct experiments on the Objects365~\cite{shao2019objects365} dataset using Deformable-DETR++~\cite{deformable} model with 900 queries.
We train the baseline model and our method for 4 epochs only.
The initial learning rate is set to 2e-4 and decays at the third epoch.
Other training settings are the same as the model trained on the COCO 2017 dataset~\cite{coco}.
As reported in~\tabref{tab:objects365}, Mr. DETR and Mr. DETR++ achieve 2.3\% and 4.5\% improvement over the baseline model in terms of AP, corroborating the effectiveness of our method.
}

{\color{black}
\subsubsection{NuImages}
We evaluate our approach on the NuImages~\cite{caesar2020nuscenes} dataset, which encompasses diverse traffic scenarios.
The experimental results are presented in~\tabref{tab:nuimages}.
Mr. DETR achieves an AP of 51.2\%, surpassing the baseline model by 2.7\%. 
Mr. DETR++ further improves performance, outperforming the baseline by 3.8\% AP, highlighting the effectiveness of our method.
These results on the NuImages dataset underscore the potential of our approach for autonomous driving applications.
}

\begin{figure}[!t]
  \centering
  \begin{subfigure}{0.24\textwidth}
    \includegraphics[width=\linewidth]{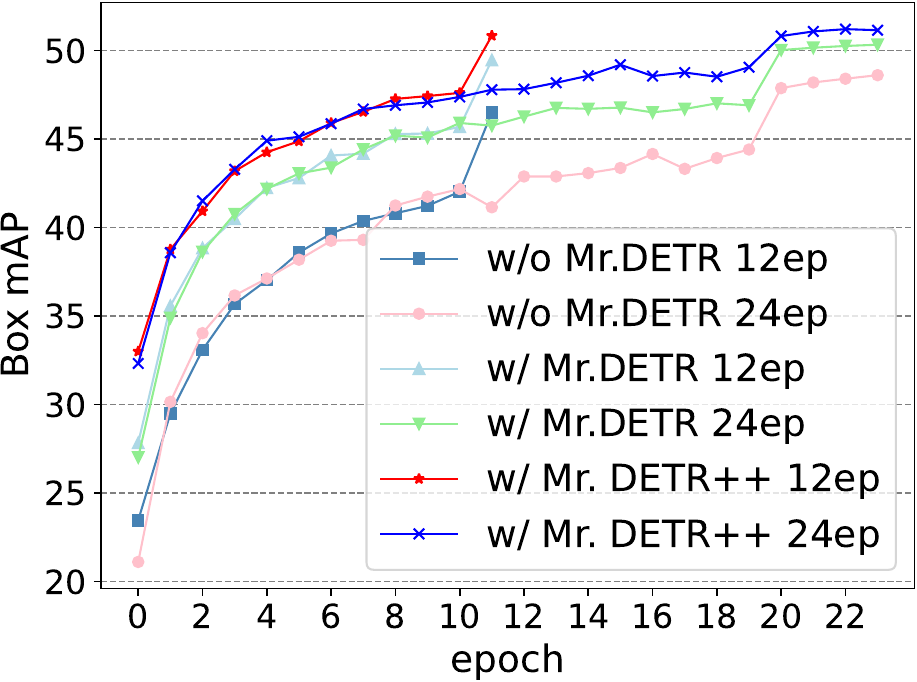}
    \caption{Evaluation results of box}\label{fig:sub1}
  \end{subfigure}
  \begin{subfigure}{0.24\textwidth}
    \includegraphics[width=\linewidth]{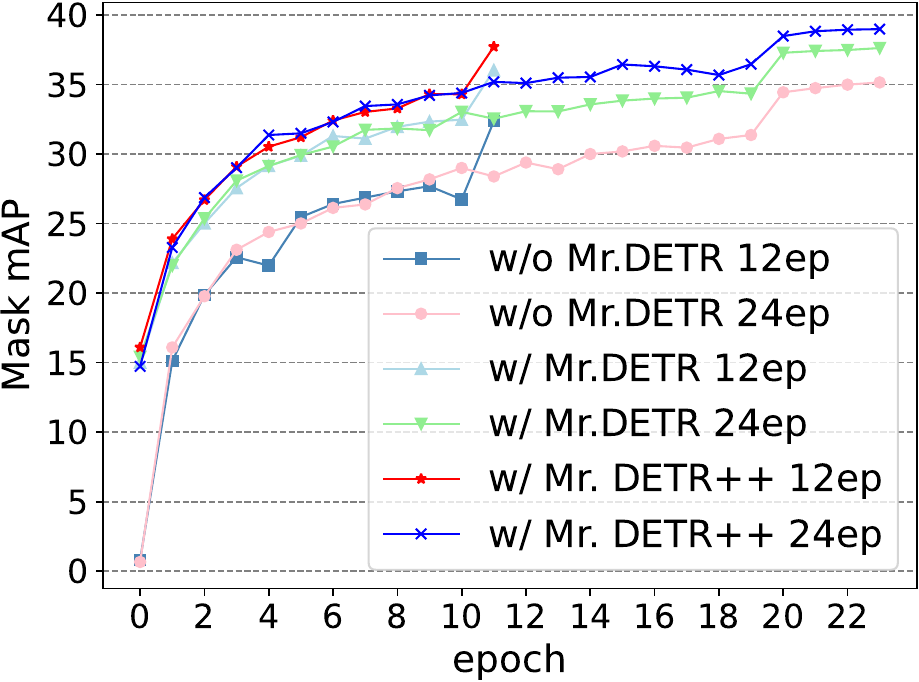}
    \caption{Evaluation results of mask}\label{fig:sub2}
  \end{subfigure}
  \caption{\textbf{Evaluation results of each epoch.} We utilize the Deformable-DETR++ (300 queries) as the baseline model, which is trained for 12 and 24 epochs, respectively.}\label{fig:trainingcurve}
  \vskip -0.2in
\end{figure}

\subsection{Instance Segmentation}
We integrate our method with the detection transformer architecture for the instance segmentation task. 
Specifically, following \cite{deformable,msdetr,zhang2025v}, we develop an object mask prediction head based on Deformable-DETR++ \cite{deformable} utilizing 300 queries. 
For simplicity, both one-to-one and one-to-many assignments consider only box and classification costs, consistent with the detection transformer. 
Also, for the IoU score learned in Mr. DETR++, we only consider the box IoU as the localization quality measure. 
We present the AP metric for both box and mask predictions in \tabref{tab:insseg}. 
Over a training schedule of 12 and 24 epochs, Mr. DETR enhances the mask AP by approximately 3.6\% and 2.5\% compared to the baseline model, respectively. 
Mr. DETR++ can improve 5.3\% and 3.9\% mask AP over Deformable-DETR++~\cite{deformable}.
These experimental results highlight the effectiveness of our method.

{\color{black}
\myPara{Convergency Curves.}
Employing Deformable-DETR++~\cite{deformable} with 300 queries, we perform training on the instance segmentation task both with and without the integration of our proposed approach.
Consistent with established methods~\cite{deformable,dino,msdetr,hu2024dac,hdetr}, models are trained using 12 and 24 epoch schedules, respectively.
The learning rate is reduced by a factor of 0.1 at the 11th and 20th epochs according to the 12 and 24 epoch schedules, respectively.
We illustrate the evaluation results for bounding box predictions in~\figref{fig:trainingcurve}(a) and for instance mask predictions in~\figref{fig:trainingcurve}(b).
The evaluation results demonstrate that our approach significantly enhances the training process of the baseline model.
}

\begin{figure*}[t]
\centering
\includegraphics[width=0.97\linewidth]{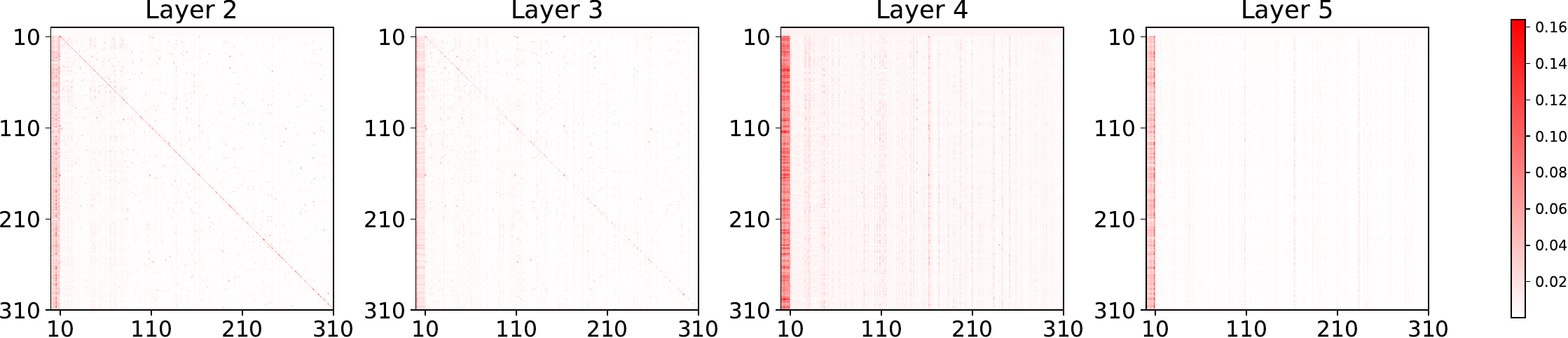}
\caption{\textbf{Visualization of attention maps for instructive self-attention.} We use Deformable-DETR++ with 300 object queries and 10 instruction tokens for this visualization. The first 10 tokens are instruction tokens. The vertical and horizontal axes represent the Query and Key, respectively. Best viewed in PDF with zoom.}
\label{fig:attnmap}
\vskip -0.1in
\end{figure*}

{\color{black}
\subsection{Panoptic Segmentation}
To further evaluate the versatility of Mr. DETR and Mr. DETR++, we extend their application to the panoptic segmentation task, which requires simultaneous prediction of bounding boxes and instance masks for `thing' classes, as well as semantic masks for `stuff' classes.
We adopt Mask DINO~\cite{li2023mask} as our baseline model.
To accommodate panoptic segmentation, we modify the one-to-many label assignment strategy in~\eqref{eq:costmatrix}, as IoU-based matching is unsuitable for semantic mask predictions. Specifically, we implement a one-to-many assignment for `stuff' classes by repeating target labels three times, determined empirically for simplicity.
Additionally, the scoring calibration is not used in panoptic segmentation.
As reported in~\tabref{tab:panoseg}, Mr. DETR achieves a 1.2\% PQ improvement over the Mask DINO baseline, while Mr. DETR++ enhances performance by 2.2\% PQ. Additionally, we provide mask AP and box AP for `thing' classes, highlighting the superior performance of our approach.
}

\begin{table*}[t]
\caption{\textbf{Panoptic segmentation results on the COCO 2017~\cite{coco} validation set.} All experiments are based on ResNet-50~\cite{he2016deep} as the backbone.}\label{tab:panoseg}
\centering
\setlength{\tabcolsep}{5pt} 
\small
\begin{tabular}{l|cc|ccc|cc}
\toprule
Model  & Epochs & Queries & PQ & PQ$^{th}$ & PQ$^{st}$ & AP$^{box}$ & AP$^{mask}$ \\
\midrule
DETR~\cite{detr} & 500 & 100 & 43.4 & 48.2 & 36.3 & - & 31.1 \\
Mask2Former~\cite{cheng2021mask2former} & 12 & 100 & 46.9 & 52.5 & 38.4 & - & 37.2 \\
Panoptic Segformer~\cite{li2022panoptic} & 12 & 353 & 48.0 & 52.3 & 41.5 & - & - \\
Mask DINO~\cite{li2023mask} & 12 & 300 & 49.0 & 54.8 & 40.2 & 43.2 & 40.4 \\
\rowcolor{blue!10}Mask DINO~\cite{li2023mask} w/ Mr. DETR  & 12 & 300 &50.2 (\textcolor{blue}{+1.2}) & 56.6 (\textcolor{blue}{+1.8}) & 40.4 (\textcolor{blue}{+0.2}) & 45.4 (\textcolor{blue}{+2.2}) & 42.0 (\textcolor{blue}{+1.6}) \\
\rowcolor{blue!10}Mask DINO~\cite{li2023mask} w/ Mr. DETR++  & 12 & 300 & \textbf{51.2} (\textcolor{blue}{+2.2}) & \textbf{57.5} (\textcolor{blue}{+2.7}) & \textbf{41.7} (\textcolor{blue}{+1.5}) & \textbf{46.0} (\textcolor{blue}{+2.8}) & \textbf{42.5} (\textcolor{blue}{+2.1})\\
\bottomrule

\end{tabular}

\vskip -0.2in
\end{table*}

\subsection{Effectiveness Analysis}

The parameters of the primary route for one-to-one prediction benefit from auxiliary route training with one-to-many predictions.
Recent works~\cite{hdetr,hu2024dac,msdetr} demonstrate that incorporating one-to-many assignment as auxiliary training enhances the localization quality of predictions.
In this section, we focus on instructive self-attention and examine its role in informing object queries for one-to-many predictions.
As illustrated in \figref{fig:attnmap}, we train Deformable-DETR++~\cite{deformable} with our proposed multi-route approach and visualize the attention map within the instructive self-attention across different decoder layers.
For visualization, we use 300 object queries and set 10 instruction tokens.
\figref{fig:attnmap} reveals that when the 300 object queries act as query and the 10 instruction tokens as key, nearly all 300 object queries exhibit strong activation with the instruction tokens.
This indicates that instruction tokens effectively convey information to object queries and subsequent network layers, aiding the model in achieving one-to-many predictions.

\subsection{Ablation Study}\label{sec:ablationstudy}
\noindent\textbf{Effectiveness of Multi-route Training in Mr. DETR.}
Our method features a primary route, referred to as Route-2 in~\figref{fig:method}, dedicated to one-to-one prediction, and two auxiliary routes, Route-1 and Route-3 for one-to-many prediction.
Route-1 utilizes shared self-attention and cross-attention with Route-2, but employs an independent FFN to achieve one-to-many prediction.
We introduce an instructive self-attention mechanism in Route-3, which shares all parameters with the primary route.
To evaluate the contribution of each auxiliary route, we perform an ablation study, as shown in~\tabref{tab:ablationcomponent}.
Route-1, with its independent FFN, enhances the baseline model by 2.0\% mAP.
Experimental results indicate that Route-3 provides a 2.8\% improvement on Deformable-DETR++~\cite{deformable}, highlighting the effectiveness of our proposed instructive self-attention.
By incorporating multi-route training, Deformable-DETR++~\cite{deformable} achieves 50.7\% mAP, surpassing the baseline by 3.1\%. 

{\color{black}
\myPara{Effectiveness of Mr. DETR++.}
As shown in Table~\ref{tab:ablationcomponent}, Mr. DETR++ replaces the auxiliary training route with two independent FFNs integrated with a route-aware MoE, yielding a 0.3\% improvement in AP.
Incorporating a scale-aware MoE into the encoder further enhances performance by 0.4\% AP.
Additionally, applying localization-aware score calibration results in a final AP of 51.8\%.
We investigate the impact of the factor $\phi$ in~\appref{appendix:localscore}.
In~\appref{appendix:detailedperformance}, we further demonstrate our model's performance across different decoder layers and routes, illustrating the effectiveness of our approach.
}

\begin{table*}[!t]
\caption{\textcolor{black}{\textbf{The ablation study of different routes in our method.}
`Route-1 (FFN)': the auxiliary training route with independent FFN. `Route-2': the primary route for one-to-one prediction. `Route-3': the auxiliary training route with instructive self-attention. `Route-1 (MoE)': the auxiliary training route with route-aware MoE. `MoE-Encoder': the scale-aware MoE in the encoder. `Calibration': the localization-aware score calibration used during inference.}}\label{tab:ablationcomponent}
\centering
\setlength{\tabcolsep}{4.5pt} 
\small
\begin{tabular}{l|ccccccccc}
\toprule
Variants & Route-1 (FFN) & Route-2 & Route-3 & Route-1 (MoE) & MoE-Encoder  & Calibration & AP & AP$_{50}$ & AP$_{75}$ \\
\midrule
& & \ding{52} & & & & &47.6 & 65.8 & 51.8\\
& \ding{52} & \ding{52} & & & & &49.6 (\textcolor{blue}{+2.0}) & 67.4 & 54.2\\
& & \ding{52} & \ding{52}& & & &50.4 (\textcolor{blue}{+2.8}) & 67.9 & 55.3 \\
\rowcolor{blue!10} Mr. DETR & \ding{52} & \ding{52} & \ding{52} & & & &\textbf{50.7} (\textcolor{blue}{+3.1}) & \textbf{68.2} & \textbf{55.4} \\
\midrule
& & \ding{52} & \ding{52} & \ding{52} & && 51.0 (\textcolor{blue}{+3.4}) & 68.7 & 56.1 \\
& & \ding{52} & \ding{52} & \ding{52} & & \ding{52} & 51.3 (\textcolor{blue}{+3.7}) & 68.5 & 55.9 \\
& & \ding{52} & \ding{52} & \ding{52} & \ding{52} & & 51.4 (\textcolor{blue}{+3.8}) & 68.8 & 56.3 \\
\rowcolor{blue!10}Mr. DETR++ & & \ding{52} & \ding{52} & \ding{52} & \ding{52} & \ding{52} & \textbf{51.8} (\textcolor{blue}{+4.2}) & \textbf{68.9} & \textbf{56.7}\\
\bottomrule

\end{tabular}

\vskip -0.1in
\end{table*}

\myPara{Designs of the Instruction Mechanism.}
\textcolor{black}{
To further reduce the trainable parameters in Route-3 and enhance parameter sharing with the primary route, we introduce the instruction mechanism.
In~\tabref{tab:ablationselfattn}, we evaluate the performance of different designs of the instruction mechanism, as discussed in~\secref{sec:insattnsec}.
Initially, we evaluate two conventional variants: (a) one with independent self-attention and (b) another without self-attention in Route-3.
Experimental results indicate that our approach surpasses these variants by 0.7\% and 0.8\% AP, respectively.
Additionally, we implement a variant (c) using independent object queries in Route-3 while sharing all other parameters with the primary route.
Our method exceeds this design by 0.8\% AP.
We also investigate an implementation (d) that integrates instruction tokens into the object queries by addition, unlike our method, which appends them.
Experimental results demonstrate the superiority of our approach, which dynamically and flexibly guides object queries to achieve one-to-many predictions.
}

\begin{table}[!t]
\caption{\textbf{Different designs of instructive self-attention.}}\label{tab:ablationselfattn}
\centering
\setlength{\tabcolsep}{3pt} 
\small
\begin{tabular}{l|cccc}
\toprule
& Variants & AP & AP$_{50}$ & AP$_{75}$\\
\midrule
(a) & Independent self-attention & 50.0 & 67.7 & 54.8 \\
(b) & Remove self-attention & 49.9 & 67.3 & 54.7 \\
(c) & Independent object queries & 49.9 & 67.5 & 54.6 \\
(d) & Instruction by addition & 50.1 & 67.6 & 54.7 \\
\rowcolor{blue!10}(e) & Instruction by concatenation (\textbf{ours}) & \textbf{50.7} & \textbf{68.2} & \textbf{55.4} \\
\bottomrule

\end{tabular}
\end{table}

\begin{table}[!t]
\caption{\textbf{The ablation study on the configurations of the route-aware MoE.} These experiments evaluate the selected scales within the scale-aware MoE, with downsampling ratios of x64, x32, x16, and x8.
}\label{tab:ablationscales}
\centering
\small
\setlength{\tabcolsep}{7pt} 
\begin{tabular}{cccc|ccccc}
\toprule
x64 & x32 & x16 & x8 & AP & AP$_{50}$ & AP$_{75}$ & GFlops \\
\midrule
& & & & 51.3 & 68.5 & 55.9 & 424.1\\
\rowcolor{blue!10}\ding{52} & & & & 51.8 & 68.9 & \textbf{56.7} & 486.8 \\
\ding{52} & \ding{52} & & & 51.7 & 68.7 & 56.4 & 503.4 \\
\ding{52} & \ding{52} & \ding{52} && \textbf{51.9} & \textbf{69.0} & \textbf{56.7} & 569.4 \\
\ding{52} & \ding{52} & \ding{52} & \ding{52} & 51.5 & 68.6 & 56.2 & 832.5 \\
\bottomrule

\end{tabular}

\vskip -0.2in
\end{table}

{\color{black}
\myPara{Ablation on the Scale-aware MoE.}
In Mr. DETR++, we extend the MoE to the transformer encoder to enhance feature representations.
However, as the transformer encoder processes a combined input sequence of image tokens from multi-scale features, directly applying MoE incurs significant computing costs.
To mitigate this, we propose a scale-aware MoE, which assigns an MoE block to low-scale tokens while sharing a single expert across all tokens.
To investigate the impact of selected low-scale tokens, we conduct an ablation study on the scales with an activated MoE block.
The results, reported in~\tabref{tab:ablationscales}, demonstrate that applying the MoE block to x64 downsampled features yields a 0.5\% AP improvement.
While using MoE on x64, x32, and x16 downsampled features achieves the highest performance of 51.9\% AP, it substantially increases computational costs.
Thus, we choose to apply MoE on x64 downsampled features, striking an optimal balance between computing efficiency and performance.

}

\begin{figure}[t]
  \centering
  \begin{subfigure}{0.48\textwidth}
    \begin{overpic}[width=1\linewidth,]
    {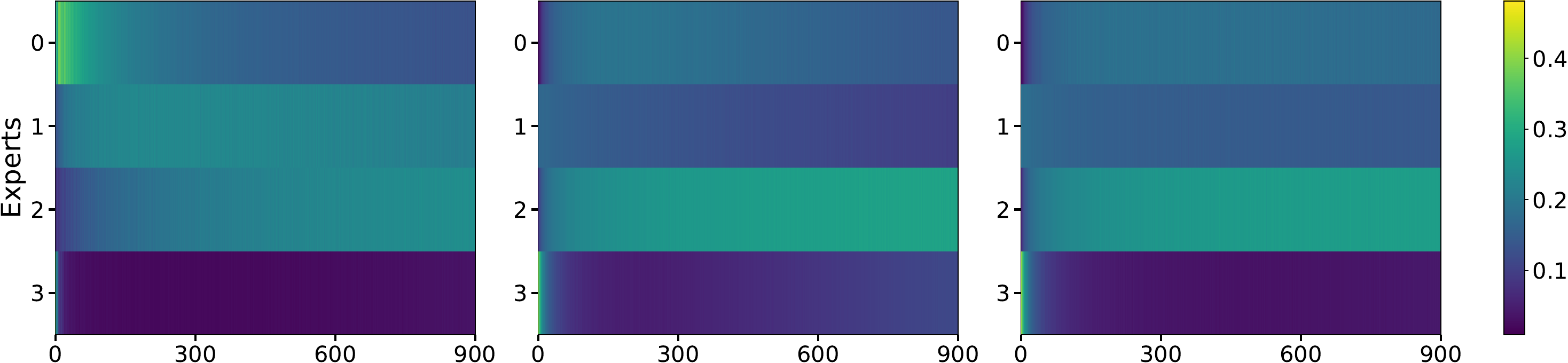}
        \put(12.5, 24){\scriptsize{Route-2}}
        \put(43, 24){\scriptsize{Route-3}}
        \put(74, 24){\scriptsize{Route-1}}
    \end{overpic}
    \caption{Layer 3 in the Decoder}
  \end{subfigure}
  \begin{subfigure}{0.48\textwidth}
    \begin{overpic}[width=1\linewidth]
    {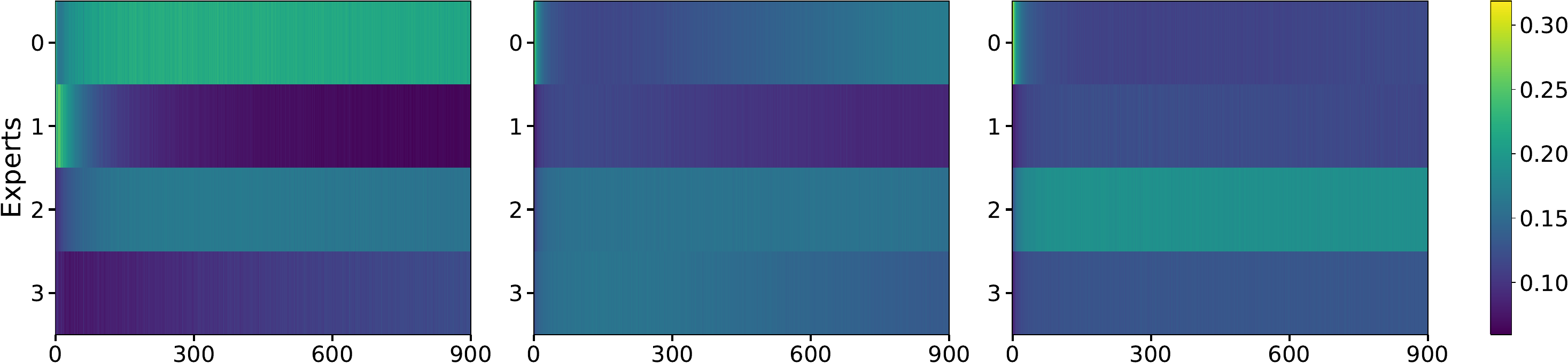}
    \end{overpic}
    \caption{Layer 4 in the Decoder}
  \end{subfigure}
  \begin{subfigure}{0.48\textwidth}
    \begin{overpic}[width=1\linewidth]
    {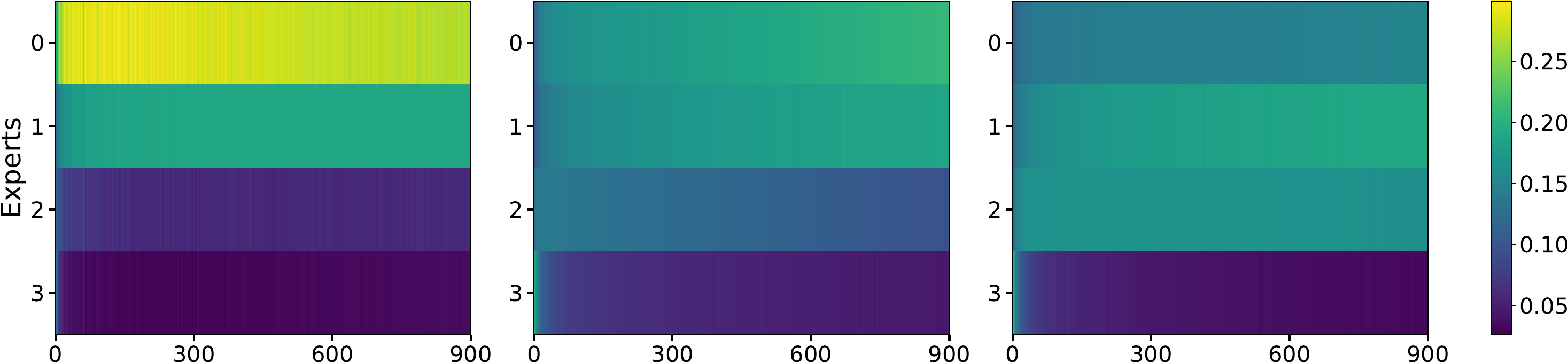}
    \end{overpic}
    \caption{Layer 5 in the Decoder}
  \end{subfigure}
  \caption{{\color{black}\textbf{Distribution of Activated Experts in the Route-aware MoE.} `Route-2': the primary route for one-to-one prediction. `Route-3': the auxiliary training route with instructive self-attention. `Route-1': the auxiliary training route with independent gating.}
  }\label{fig:activatedexperts}
  \vskip -0.2in
\end{figure}

{\color{black}
\myPara{Distribution of Activated Experts in the Route-aware MoE.}
To save computing cost, we use a route-aware MoE in the last three layers of the transformer decoder.
As shown in~\figref{fig:activatedexperts}, we visualize the distribution of activated experts in each gating function.
In layers 3 and 4, overall, all routes tend to prefer three experts, but the two auxiliary training routes exhibit preference patterns distinct from Route-2.
For instance, in layer 3, Route-2’s gating shows a high activation for Expert-0, whereas Route-3 and Route-1 do not.
However, Route-2’s expert selection still significantly overlaps with that of the other routes; for example, Expert-2 is consistently favored across all three routes.
In the final layer, Route-2 demonstrates a strong preference for the first two experts, while the other two routes distribute their preferences more evenly across the first three experts.
This suggests that a route-aware MoE can both avoid task conflicts through expert selection and facilitate knowledge sharing.
}

\myPara{Training Cost.}
We measure the training time of different methods in~\tabref{tab:traintime}.
The training time denotes the average duration of each epoch.
We evaluate training time on 8 NVIDIA 3090 GPUs with a batch size of 16.
The experimental results indicate that our proposed method achieves an effective trade-off between performance and training costs.

\begin{table}[t]
    \centering
    \small
    \setlength{\tabcolsep}{9pt}
    \caption{\textbf{Comparison of training time (minutes) of various methods.} All methods utilize Deformable-DETR++ with 300 queries as the baseline. The training time represents the average duration (minutes) per training epoch.} 
    \label{tab:traintime}
    \begin{tabular}{l|c|c}
    \toprule
    Model & Duration & AP \\
    \midrule
    Deformable-DETR++~\cite{deformable} & 84   & 47.0 \\
    H-DETR~\cite{hdetr} &   104  (\textcolor{red}{+20})  & 48.7 (\textcolor{blue}{+1.7}) \\
    MS-DETR~\cite{msdetr} & 96 (\textcolor{red}{+12}) & 48.8 (\textcolor{blue}{+1.8})\\
    Mr. DETR (ours)  & 101 (\textcolor{red}{+17}) 
    & 49.5 (\textcolor{blue}{+2.5})\\
    Mr. DETR++ (ours) & 127 (\textcolor{red}{+43}) & \textbf{51.0}  (\textcolor{blue}{+4.0})\\
    \bottomrule
    \end{tabular}
    \vskip -0.2in

\end{table}

{\color{black}
\section{Mechanism Analysis}
\noindent\textbf{Does the non-shared component or instructive self-attention disrupt the necessary information for one-to-one prediction in object queries?}
We observe that the one-to-many route equipped with a non-shared component can still perform well, as discussed in~\secref{sec:introduction}.
To investigate whether Route-1 (FFN) and Route-3 (instructive self-attention) achieve one-to-many prediction by discarding essential information for one-to-one prediction in object queries, we conduct probing experiments using Mr. DETR as an example.
Specifically, we freeze all parameters of the pre-trained Mr. DETR model and attach a trainable classification layer with random initialization to each route.
These trainable layers are then trained under a one-to-one assignment to extract any remaining information relevant to one-to-one prediction from the object queries.
The results of the probing experiment, as shown in~\tabref{tab:probing}, indicate that the probe layer for Route-2 achieves comparable performance (50.3 AP) to the oracle one-to-one prediction.
In contrast, the probe layers for Route-1 and Route-3 perform significantly worse, both reaching only about 20.0 AP, far below the performance when equipping NMS (49.0 AP).
This suggests that while the probe layer for Route-2 can extract sufficient information for one-to-one prediction, the probe layers for Route-1 and Route-3 fail to do so.
In summary, the experimental results demonstrate that the non-shared component and instructive self-attention significantly disrupt the information necessary for one-to-one prediction in object queries.

\begin{table}[t]
    \centering
    \small
    \setlength{\tabcolsep}{7pt}
    \caption{{\color{black}\textbf{Probing Analysis.} `Route-2': the primary route for one-to-one prediction. `Route-1': the auxiliary training route with independent FFN. `Route-3': the auxiliary training route with the instructive self-attention.}}
    \begin{tabular}{l|c|c}
    \toprule
    Model & w/o NMS & w/ NMS \\
    \midrule
    Deformable-DETR++~\cite{deformable} & 47.6 & - \\
    \midrule
    Mr. DETR Route-2 & 50.7 & 50.8 \\
    Mr. DETR Route-2 + Probing & 50.3 & 50.5 \\
    Mr. DETR Route-1 & 12.8 & 51.0 \\
    Mr. DETR Route-1 + Probing & 19.3 & 49.0 \\
    Mr. DETR Route-3 & 12.8 & 51.1 \\
    Mr. DETR Route-3 + Probing & 20.8 & 48.9 \\
    \midrule
    Training with o2m only & 13.3 & 50.2\\
    + Probing for o2o & 13.3  & 50.1 \\
    \bottomrule
    \end{tabular}
    
    \label{tab:probing}
    \vskip -0.2in
\end{table}

\noindent\myPara{Can any non-shared component in the decoder avoid conflict between one-to-one and one-to-many prediction?}
As discussed in~\secref{sec:introduction}, we find that any non-shared component can mitigate the conflict between one-to-one and one-to-many prediction while benefiting the primary one-to-one prediction.
To further investigate the mechanism underlying multi-route training, we conduct an analysis experiment, as shown in~\tabref{tab:probing}.
Specifically, we first train an oracle model using only one-to-many supervision and then perform probing analysis on the pre-trained model.
The experimental results reveal that object queries supervised solely by one-to-many assignment lack the necessary information for one-to-one prediction.
In conjunction with the earlier conclusion that non-shared components disrupt the information required for one-to-one prediction, we infer that routes trained without one-to-one supervision cannot retain the information essential for one-to-one prediction.
Moreover, any non-shared component inherently disrupts this information, thereby preventing conflict between one-to-one and one-to-many predictions.
In summary, these non-shared components play a key role in avoiding conflicts between the two types of prediction.
}

\section{Conclusion}
Regarding the model with auxiliary one-to-many training as a multi-task framework, we examine the roles of each component in the transformer decoder for two training targets.
Our empirical findings indicate that any independent component in the decoder can effectively learn both one-to-one and one-to-many training targets, even when other components are shared.
Building on this insight, we propose a multi-route training mechanism, featuring a primary route and two auxiliary training routes.
\textcolor{black}{
The first auxiliary route incorporates our proposed instructive self-attention, which dynamically and flexibly guides object queries for one-to-many prediction.
The second auxiliary route is enhanced by our proposed route-aware mixture-of-experts, enabling knowledge sharing while mitigating potential conflicts between routes.
Additionally, the transformer encoder is further enhanced with our proposed scale-aware MoE, applied to low-scale features to reduce computational cost.
Notably, the auxiliary training routes are discarded during inference.
Extensive experiments across multiple tasks, including object detection, instance segmentation, and panoptic segmentation, validate the effectiveness and versatility of our approach.
}

\section*{Acknowledgement}
This work is supported by National Natural Science Foundation of China (Grant No. 62306251), Hong Kong Research Grant Council - Early Career Scheme (Grant No. 27208022), 
and HKU Seed Fund for Basic Research.
The computations were performed partly using research computing facilities offered by Information Technology Services, The University of Hong Kong.

\bibliographystyle{IEEEtran}
\bibliography{ref}

\begin{thebibliography}{10}
\providecommand{\url}[1]{#1}
\csname url@samestyle\endcsname
\providecommand{\newblock}{\relax}
\providecommand{\bibinfo}[2]{#2}
\providecommand{\BIBentrySTDinterwordspacing}{\spaceskip=0pt\relax}
\providecommand{\BIBentryALTinterwordstretchfactor}{4}
\providecommand{\BIBentryALTinterwordspacing}{\spaceskip=\fontdimen2\font plus
\BIBentryALTinterwordstretchfactor\fontdimen3\font minus \fontdimen4\font\relax}
\providecommand{\BIBforeignlanguage}[2]{{%
\expandafter\ifx\csname l@#1\endcsname\relax
\typeout{** WARNING: IEEEtran.bst: No hyphenation pattern has been}%
\typeout{** loaded for the language `#1'. Using the pattern for}%
\typeout{** the default language instead.}%
\else
\language=\csname l@#1\endcsname
\fi
#2}}
\providecommand{\BIBdecl}{\relax}
\BIBdecl

\bibitem{detr}
N.~Carion, F.~Massa, G.~Synnaeve, N.~Usunier, A.~Kirillov, and S.~Zagoruyko, ``End-to-end object detection with transformers,'' in \emph{Eur. Conf. Comput. Vis.}, 2020.

\bibitem{liu2022dab}
S.~Liu, F.~Li, H.~Zhang, X.~Yang, X.~Qi, H.~Su, J.~Zhu, and L.~Zhang, ``Dab-detr: Dynamic anchor boxes are better queries for detr,'' in \emph{Int. Conf. Learn. Represent.}, 2022.

\bibitem{meng2021conditional}
D.~Meng, X.~Chen, Z.~Fan, G.~Zeng, H.~Li, Y.~Yuan, L.~Sun, and J.~Wang, ``Conditional detr for fast training convergence,'' in \emph{Int. Conf. Comput. Vis.}, 2021.

\bibitem{wang2022anchor}
Y.~Wang, X.~Zhang, T.~Yang, and J.~Sun, ``Anchor detr: Query design for transformer-based detector,'' in \emph{AAAI Conf. Artif. Intell.}, 2022.

\bibitem{ren2016faster}
S.~Ren, K.~He, R.~Girshick, and J.~Sun, ``Faster r-cnn: Towards real-time object detection with region proposal networks,'' \emph{IEEE Trans. Pattern Anal. Mach. Intell.}, 2016.

\bibitem{ross2017focal}
T.-Y. Lin, P.~Goyal, R.~B. Girshick, K.~He, and P.~Doll{\'a}r, ``Focal loss for dense object detection,'' \emph{Int. Conf. Comput. Vis.}, 2017.

\bibitem{tian2020fcos}
Z.~Tian, C.~Shen, H.~Chen, and T.~He, ``Fcos: A simple and strong anchor-free object detector,'' \emph{IEEE Trans. Pattern Anal. Mach. Intell.}, 2020.

\bibitem{hdetr}
D.~Jia, Y.~Yuan, H.~He, X.~Wu, H.~Yu, W.~Lin, L.~Sun, C.~Zhang, and H.~Hu, ``Detrs with hybrid matching,'' in \emph{IEEE Conf. Comput. Vis. Pattern Recog.}, 2023.

\bibitem{dino}
H.~Zhang, F.~Li, S.~Liu, L.~Zhang, H.~Su, J.~Zhu, L.~Ni, and H.-Y. Shum, ``Dino: Detr with improved denoising anchor boxes for end-to-end object detection,'' in \emph{Int. Conf. Learn. Represent.}, 2023.

\bibitem{sun2021rethinking}
Z.~Sun, S.~Cao, Y.~Yang, and K.~M. Kitani, ``Rethinking transformer-based set prediction for object detection,'' in \emph{Int. Conf. Comput. Vis.}, 2021.

\bibitem{deformable}
X.~Zhu, W.~Su, L.~Lu, B.~Li, X.~Wang, and J.~Dai, ``Deformable detr: Deformable transformers for end-to-end object detection,'' in \emph{Int. Conf. Learn. Represent.}, 2021.

\bibitem{cai2023align}
Z.~Cai, S.~Liu, G.~Wang, Z.~Ge, X.~Zhang, and D.~Huang, ``Align-detr: Improving detr with simple iou-aware bce loss,'' in \emph{Brit. Mach. Vis. Conf.}, 2024.

\bibitem{hu2024dac}
Z.~Hu, Y.~Sun, J.~Wang, and Y.~Yang, ``Dac-detr: Divide the attention layers and conquer,'' \emph{Adv. Neural Inform. Process. Syst.}, 2024.

\bibitem{msdetr}
C.~Zhao, Y.~Sun, W.~Wang, Q.~Chen, E.~Ding, Y.~Yang, and J.~Wang, ``Ms-detr: Efficient detr training with mixed supervision,'' in \emph{IEEE Conf. Comput. Vis. Pattern Recog.}, 2024.

\bibitem{li2022dn}
F.~Li, H.~Zhang, S.~Liu, J.~Guo, L.~M. Ni, and L.~Zhang, ``Dn-detr: Accelerate detr training by introducing query denoising,'' \emph{IEEE Trans. Pattern Anal. Mach. Intell.}, 2024.

\bibitem{chen2023group}
Q.~Chen, X.~Chen, J.~Wang, S.~Zhang, K.~Yao, H.~Feng, J.~Han, E.~Ding, G.~Zeng, and J.~Wang, ``Group detr: Fast detr training with group-wise one-to-many assignment,'' in \emph{Int. Conf. Comput. Vis.}, 2023.

\bibitem{ouyang2022nms}
J.~Ouyang-Zhang, J.~H. Cho, X.~Zhou, and P.~Kr{\"a}henb{\"u}hl, ``Nms strikes back,'' \emph{arXiv preprint arXiv:2212.06137}, 2022.

\bibitem{zhang2021varifocalnet}
H.~Zhang, Y.~Wang, F.~Dayoub, and N.~Sunderhauf, ``Varifocalnet: An iou-aware dense object detector,'' in \emph{IEEE Conf. Comput. Vis. Pattern Recog.}, 2021.

\bibitem{liu2023detection}
S.~Liu, T.~Ren, J.~Chen, Z.~Zeng, H.~Zhang, F.~Li, H.~Li, J.~Huang, H.~Su, J.~Zhu \emph{et~al.}, ``Detection transformer with stable matching,'' in \emph{Int. Conf. Comput. Vis.}, 2023.

\bibitem{pu2024rank}
Y.~Pu, W.~Liang, Y.~Hao, Y.~Yuan, Y.~Yang, C.~Zhang, H.~Hu, and G.~Huang, ``Rank-detr for high quality object detection,'' \emph{Adv. Neural Inform. Process. Syst.}, 2024.

\bibitem{zhang2025mr}
C.-B. Zhang, Y.~Zhong, and K.~Han, ``Mr. detr: Instructive multi-route training for detection transformers,'' in \emph{IEEE Conf. Comput. Vis. Pattern Recog.}, 2025.

\bibitem{shao2019objects365}
S.~Shao, Z.~Li, T.~Zhang, C.~Peng, G.~Yu, X.~Zhang, J.~Li, and J.~Sun, ``Objects365: A large-scale, high-quality dataset for object detection,'' in \emph{Int. Conf. Comput. Vis.}, 2019.

\bibitem{caesar2020nuscenes}
H.~Caesar, V.~Bankiti, A.~H. Lang, S.~Vora, V.~E. Liong, Q.~Xu, A.~Krishnan, Y.~Pan, G.~Baldan, and O.~Beijbom, ``nuscenes: A multimodal dataset for autonomous driving,'' in \emph{IEEE Conf. Comput. Vis. Pattern Recog.}, 2020.

\bibitem{zheng2021yolox}
G.~Zheng, L.~Songtao, W.~Feng, L.~Zeming, S.~Jian \emph{et~al.}, ``Yolox: Exceeding yolo series in 2021,'' \emph{arXiv preprint arXiv:2107.08430}, 2021.

\bibitem{jin2022you}
Z.~Jin, D.~Yu, L.~Song, Z.~Yuan, and L.~Yu, ``You should look at all objects,'' in \emph{Eur. Conf. Comput. Vis.}, 2022.

\bibitem{dai2021dynamic}
X.~Dai, Y.~Chen, J.~Yang, P.~Zhang, L.~Yuan, and L.~Zhang, ``Dynamic detr: End-to-end object detection with dynamic attention,'' in \emph{Int. Conf. Comput. Vis.}, 2021.

\bibitem{gao2021fast}
P.~Gao, M.~Zheng, X.~Wang, J.~Dai, and H.~Li, ``Fast convergence of detr with spatially modulated co-attention,'' in \emph{Int. Conf. Comput. Vis.}, 2021.

\bibitem{ye2023cascade}
M.~Ye, L.~Ke, S.~Li, Y.-W. Tai, C.-K. Tang, M.~Danelljan, and F.~Yu, ``Cascade-detr: delving into high-quality universal object detection,'' in \emph{Int. Conf. Comput. Vis.}, 2023.

\bibitem{gao2024ease}
Y.~Gao, Y.~Sun, X.~Ding, C.~Zhao, and S.~Liu, ``Ease-detr: Easing the competition among object queries,'' in \emph{IEEE Conf. Comput. Vis. Pattern Recog.}, 2024.

\bibitem{hou2024relation}
X.~Hou, M.~Liu, S.~Zhang, P.~Wei, B.~Chen, and X.~Lan, ``Relation detr: Exploring explicit position relation prior for object detection,'' in \emph{Eur. Conf. Comput. Vis.}, 2024.

\bibitem{liu2023sap}
Y.~Liu, Y.~Zhang, Y.~Wang, Y.~Zhang, J.~Tian, Z.~Shi, J.~Fan, and Z.~He, ``Sap-detr: bridging the gap between salient points and queries-based transformer detector for fast model convergency,'' in \emph{IEEE Conf. Comput. Vis. Pattern Recog.}, 2023.

\bibitem{zhang2022accelerating}
G.~Zhang, Z.~Luo, Y.~Yu, K.~Cui, and S.~Lu, ``Accelerating detr convergence via semantic-aligned matching,'' in \emph{IEEE Conf. Comput. Vis. Pattern Recog.}, 2022.

\bibitem{zhao2024hybrid}
J.~Zhao, F.~Wei, and C.~Xu, ``Hybrid proposal refiner: Revisiting detr series from the faster r-cnn perspective,'' in \emph{IEEE Conf. Comput. Vis. Pattern Recog.}, 2024.

\bibitem{nan2025mi}
Z.~Nan, X.~Li, J.~Dai, and T.~Xiang, ``Mi-detr: An object detection model with multi-time inquiries mechanism,'' \emph{arXiv preprint arXiv:2503.01463}, 2025.

\bibitem{yao2021efficient}
Z.~Yao, J.~Ai, B.~Li, and C.~Zhang, ``Efficient detr: improving end-to-end object detector with dense prior,'' \emph{arXiv preprint arXiv:2104.01318}, 2021.

\bibitem{hou2024salience}
X.~Hou, M.~Liu, S.~Zhang, P.~Wei, and B.~Chen, ``Salience detr: Enhancing detection transformer with hierarchical salience filtering refinement,'' in \emph{IEEE Conf. Comput. Vis. Pattern Recog.}, 2024.

\bibitem{zhang2023dense}
S.~Zhang, X.~Wang, J.~Wang, J.~Pang, C.~Lyu, W.~Zhang, P.~Luo, and K.~Chen, ``Dense distinct query for end-to-end object detection,'' in \emph{IEEE Conf. Comput. Vis. Pattern Recog.}, 2023.

\bibitem{roh2021sparse}
B.~Roh, J.~Shin, W.~Shin, and S.~Kim, ``Sparse detr: Efficient end-to-end object detection with learnable sparsity,'' in \emph{Int. Conf. Learn. Represent.}, 2022.

\bibitem{lin2022d}
J.~Lin, X.~Mao, Y.~Chen, L.~Xu, Y.~He, and H.~Xue, ``D\^{} 2etr: Decoder-only detr with computationally efficient cross-scale attention,'' \emph{arXiv preprint arXiv:2203.00860}, 2022.

\bibitem{zhao2024detrs}
Y.~Zhao, W.~Lv, S.~Xu, J.~Wei, G.~Wang, Q.~Dang, Y.~Liu, and J.~Chen, ``Detrs beat yolos on real-time object detection,'' in \emph{IEEE Conf. Comput. Vis. Pattern Recog.}, 2024.

\bibitem{li2023lite}
F.~Li, A.~Zeng, S.~Liu, H.~Zhang, H.~Li, L.~Zhang, and L.~M. Ni, ``Lite detr: An interleaved multi-scale encoder for efficient detr,'' in \emph{IEEE Conf. Comput. Vis. Pattern Recog.}, 2023.

\bibitem{chen2024lw}
Q.~Chen, X.~Su, X.~Zhang, J.~Wang, J.~Chen, Y.~Shen, C.~Han, Z.~Chen, W.~Xu, F.~Li \emph{et~al.}, ``Lw-detr: A transformer replacement to yolo for real-time detection,'' \emph{arXiv preprint arXiv:2406.03459}, 2024.

\bibitem{zhang2023decoupled}
M.~Zhang, G.~Song, Y.~Liu, and H.~Li, ``Decoupled detr: Spatially disentangling localization and classification for improved end-to-end object detection,'' in \emph{Int. Conf. Comput. Vis.}, 2023.

\bibitem{zhang2020bridging}
S.~Zhang, C.~Chi, Y.~Yao, Z.~Lei, and S.~Z. Li, ``Bridging the gap between anchor-based and anchor-free detection via adaptive training sample selection,'' in \emph{IEEE Conf. Comput. Vis. Pattern Recog.}, 2020.

\bibitem{ge2021ota}
Z.~Ge, S.~Liu, Z.~Li, O.~Yoshie, and J.~Sun, ``Ota: Optimal transport assignment for object detection,'' in \emph{IEEE Conf. Comput. Vis. Pattern Recog.}, 2021.

\bibitem{feng2021tood}
C.~Feng, Y.~Zhong, Y.~Gao, M.~R. Scott, and W.~Huang, ``Tood: Task-aligned one-stage object detection,'' in \emph{Int. Conf. Comput. Vis.}, 2021.

\bibitem{gao2022adamixer}
Z.~Gao, L.~Wang, B.~Han, and S.~Guo, ``Adamixer: A fast-converging query-based object detector,'' in \emph{IEEE Conf. Comput. Vis. Pattern Recog.}, 2022.

\bibitem{chen2022recurrent}
Z.~Chen, J.~Zhang, and D.~Tao, ``Recurrent glimpse-based decoder for detection with transformer,'' in \emph{IEEE Conf. Comput. Vis. Pattern Recog.}, 2022.

\bibitem{yang2022querydet}
C.~Yang, Z.~Huang, and N.~Wang, ``Querydet: Cascaded sparse query for accelerating high-resolution small object detection,'' in \emph{IEEE Conf. Comput. Vis. Pattern Recog.}, 2022.

\bibitem{li2022exploring}
Y.~Li, H.~Mao, R.~Girshick, and K.~He, ``Exploring plain vision transformer backbones for object detection,'' in \emph{Eur. Conf. Comput. Vis.}, 2022.

\bibitem{cao2022cf}
X.~Cao, P.~Yuan, B.~Feng, and K.~Niu, ``Cf-detr: Coarse-to-fine transformers for end-to-end object detection,'' in \emph{AAAI Conf. Artif. Intell.}, 2022.

\bibitem{fang2024feataug}
R.~Fang, P.~Gao, A.~Zhou, Y.~Cai, S.~Liu, J.~Dai, and H.~Li, ``Feataug-detr: Enriching one-to-many matching for detrs with feature augmentation,'' \emph{IEEE Trans. Pattern Anal. Mach. Intell.}, 2024.

\bibitem{kuhn1955hungarian}
H.~W. Kuhn, ``The hungarian method for the assignment problem,'' \emph{Naval Res. Logist.}, 1955.

\bibitem{cui2023learning}
Y.~Cui, L.~Yang, and H.~Yu, ``Learning dynamic query combinations for transformer-based object detection and segmentation,'' in \emph{Int. Conf. Mach. Learn.}, 2023.

\bibitem{teng2023stageinteractor}
Y.~Teng, H.~Liu, S.~Guo, and L.~Wang, ``Stageinteractor: Query-based object detector with cross-stage interaction,'' in \emph{Int. Conf. Comput. Vis.}, 2023.

\bibitem{chen2023enhanced}
F.~Chen, H.~Zhang, K.~Hu, Y.-K. Huang, C.~Zhu, and M.~Savvides, ``Enhanced training of query-based object detection via selective query recollection,'' in \emph{IEEE Conf. Comput. Vis. Pattern Recog.}, 2023.

\bibitem{zong2023detrs}
Z.~Zong, G.~Song, and Y.~Liu, ``Detrs with collaborative hybrid assignments training,'' in \emph{Int. Conf. Comput. Vis.}, 2023.

\bibitem{wang2024kd}
Y.~Wang, X.~Li, S.~Weng, G.~Zhang, H.~Yue, H.~Feng, J.~Han, and E.~Ding, ``Kd-detr: Knowledge distillation for detection transformer with consistent distillation points sampling,'' in \emph{IEEE Conf. Comput. Vis. Pattern Recog.}, 2024.

\bibitem{chang2023detrdistill}
J.~Chang, S.~Wang, H.-M. Xu, Z.~Chen, C.~Yang, and F.~Zhao, ``Detrdistill: A universal knowledge distillation framework for detr-families,'' in \emph{Int. Conf. Comput. Vis.}, 2023.

\bibitem{huang2023teach}
L.~Huang, K.~Lu, G.~Song, L.~Wang, S.~Liu, Y.~Liu, and H.~Li, ``Teach-detr: Better training detr with teachers,'' \emph{IEEE Trans. Pattern Anal. Mach. Intell.}, 2023.

\bibitem{hinton2015distilling}
G.~Hinton, O.~Vinyals, and J.~Dean, ``Distilling the knowledge in a neural network,'' \emph{arXiv preprint arXiv:1503.02531}, 2015.

\bibitem{zhang2021delving}
C.-B. Zhang, P.-T. Jiang, Q.~Hou, Y.~Wei, Q.~Han, Z.~Li, and M.-M. Cheng, ``Delving deep into label smoothing,'' \emph{IEEE Trans. Image Process.}, 2021.

\bibitem{jacobs1991adaptive}
R.~A. Jacobs, M.~I. Jordan, S.~J. Nowlan, and G.~E. Hinton, ``Adaptive mixtures of local experts,'' \emph{Neural computation}, 1991.

\bibitem{shazeer2017outrageously}
N.~Shazeer, A.~Mirhoseini, K.~Maziarz, A.~Davis, Q.~Le, G.~Hinton, and J.~Dean, ``Outrageously large neural networks: The sparsely-gated mixture-of-experts layer,'' in \emph{Int. Conf. Learn. Represent.}, 2017.

\bibitem{fedus2022switch}
W.~Fedus, B.~Zoph, and N.~Shazeer, ``Switch transformers: Scaling to trillion parameter models with simple and efficient sparsity,'' \emph{J. Machine Learn. Research}, 2022.

\bibitem{liu2024deepseek}
A.~Liu, B.~Feng, B.~Xue, B.~Wang, B.~Wu, C.~Lu, C.~Zhao, C.~Deng, C.~Zhang, C.~Ruan \emph{et~al.}, ``Deepseek-v3 technical report,'' \emph{arXiv preprint arXiv:2412.19437}, 2024.

\bibitem{riquelme2021scaling}
C.~Riquelme, J.~Puigcerver, B.~Mustafa, M.~Neumann, R.~Jenatton, A.~Susano~Pinto, D.~Keysers, and N.~Houlsby, ``Scaling vision with sparse mixture of experts,'' in \emph{Adv. Neural Inform. Process. Syst.}, 2021.

\bibitem{renggli2022learning}
C.~Renggli, A.~S. Pinto, N.~Houlsby, B.~Mustafa, J.~Puigcerver, and C.~Riquelme, ``Learning to merge tokens in vision transformers,'' \emph{arXiv preprint arXiv:2202.12015}, 2022.

\bibitem{wu2022residual}
L.~Wu, M.~Liu, Y.~Chen, D.~Chen, X.~Dai, and L.~Yuan, ``Residual mixture of experts,'' \emph{arXiv preprint arXiv:2204.09636}, 2022.

\bibitem{zhang2023robust}
Y.~Zhang, R.~Cai, T.~Chen, G.~Zhang, H.~Zhang, P.-Y. Chen, S.~Chang, Z.~Wang, and S.~Liu, ``Robust mixture-of-expert training for convolutional neural networks,'' in \emph{Int. Conf. Comput. Vis.}, 2023.

\bibitem{fei2024scaling}
Z.~Fei, M.~Fan, C.~Yu, D.~Li, and J.~Huang, ``Scaling diffusion transformers to 16 billion parameters,'' \emph{arXiv preprint arXiv:2407.11633}, 2024.

\bibitem{fan2022m3vit}
Z.~Fan, R.~Sarkar, Z.~Jiang, T.~Chen, K.~Zou, Y.~Cheng, C.~Hao, Z.~Wang \emph{et~al.}, ``M$^3$vit: Mixture-of-experts vision transformer for efficient multi-task learning with model-accelerator co-design,'' in \emph{Adv. Neural Inform. Process. Syst.}, 2022.

\bibitem{chen2023mod}
Z.~Chen, Y.~Shen, M.~Ding, Z.~Chen, H.~Zhao, E.~G. Learned-Miller, and C.~Gan, ``Mod-squad: Designing mixtures of experts as modular multi-task learners,'' in \emph{IEEE Conf. Comput. Vis. Pattern Recog.}, 2023.

\bibitem{chen2023adamv}
T.~Chen, X.~Chen, X.~Du, A.~Rashwan, F.~Yang, H.~Chen, Z.~Wang, and Y.~Li, ``Adamv-moe: Adaptive multi-task vision mixture-of-experts,'' in \emph{Int. Conf. Comput. Vis.}, 2023.

\bibitem{yang2024multi}
Y.~Yang, P.-T. Jiang, Q.~Hou, H.~Zhang, J.~Chen, and B.~Li, ``Multi-task dense prediction via mixture of low-rank experts,'' in \emph{IEEE Conf. Comput. Vis. Pattern Recog.}, 2024.

\bibitem{hu2022lora}
E.~J. Hu, Y.~Shen, P.~Wallis, Z.~Allen-Zhu, Y.~Li, S.~Wang, L.~Wang, W.~Chen \emph{et~al.}, ``Lora: Low-rank adaptation of large language models,'' \emph{ICLR}, 2022.

\bibitem{brown2020language}
B.~Mann, N.~Ryder, M.~Subbiah, J.~Kaplan, P.~Dhariwal, A.~Neelakantan, P.~Shyam, G.~Sastry, A.~Askell, S.~Agarwal \emph{et~al.}, ``Language models are few-shot learners,'' \emph{arXiv preprint arXiv:2005.14165}, 2020.

\bibitem{shin2020autoprompt}
T.~Shin, Y.~Razeghi, R.~L. Logan~IV, E.~Wallace, and S.~Singh, ``Autoprompt: Eliciting knowledge from language models with automatically generated prompts,'' \emph{arXiv preprint arXiv:2010.15980}, 2020.

\bibitem{jiang2020can}
Z.~Jiang, F.~F. Xu, J.~Araki, and G.~Neubig, ``How can we know what language models know?'' \emph{Trans. Assoc. Comput. Linguist.}, 2020.

\bibitem{li2021prefix}
X.~L. Li and P.~Liang, ``Prefix-tuning: Optimizing continuous prompts for generation,'' in \emph{Annu. Meet. Assoc. Comput. Linguist.}, 2021.

\bibitem{lester2021power}
B.~Lester, R.~Al-Rfou, and N.~Constant, ``The power of scale for parameter-efficient prompt tuning,'' in \emph{Conf. Empir. Methods Nat. Lang. Process.}, 2021.

\bibitem{liu2021p}
X.~Liu, K.~Ji, Y.~Fu, W.~L. Tam, Z.~Du, Z.~Yang, and J.~Tang, ``P-tuning v2: Prompt tuning can be comparable to fine-tuning universally across scales and tasks,'' in \emph{Annu. Meet. Assoc. Comput. Linguist.}, 2022.

\bibitem{zhou2022learning}
K.~Zhou, J.~Yang, C.~C. Loy, and Z.~Liu, ``Learning to prompt for vision-language models,'' \emph{Int. J. Comput. Vis.}, 2022.

\bibitem{ju2022prompting}
C.~Ju, T.~Han, K.~Zheng, Y.~Zhang, and W.~Xie, ``Prompting visual-language models for efficient video understanding,'' in \emph{Eur. Conf. Comput. Vis.}, 2022.

\bibitem{wang2022learning}
Z.~Wang, Z.~Zhang, C.-Y. Lee, H.~Zhang, R.~Sun, X.~Ren, G.~Su, V.~Perot, J.~Dy, and T.~Pfister, ``Learning to prompt for continual learning,'' in \emph{IEEE Conf. Comput. Vis. Pattern Recog.}, 2022.

\bibitem{li2025imove}
J.~Li, Y.~Shi, Z.~Ma, H.~Xu, F.~Cheng, H.~Xiao, R.~Kang, F.~Yang, T.~Gao, and D.~Zhang, ``imove: Instance-motion-aware video understanding,'' \emph{arXiv preprint arXiv:2502.11594}, 2025.

\bibitem{jia2022visual}
M.~Jia, L.~Tang, B.-C. Chen, C.~Cardie, S.~Belongie, B.~Hariharan, and S.-N. Lim, ``Visual prompt tuning,'' in \emph{Eur. Conf. Comput. Vis.}, 2022.

\bibitem{shen2024multitask}
S.~Shen, S.~Yang, T.~Zhang, B.~Zhai, J.~E. Gonzalez, K.~Keutzer, and T.~Darrell, ``Multitask vision-language prompt tuning,'' in \emph{IEEE Winter Conf. Appl. Comput. Vis.}, 2024.

\bibitem{yoo2023improving}
S.~Yoo, E.~Kim, D.~Jung, J.~Lee, and S.~Yoon, ``Improving visual prompt tuning for self-supervised vision transformers,'' in \emph{Int. Conf. Mach. Learn.}, 2023.

\bibitem{coco}
T.-Y. Lin, M.~Maire, S.~Belongie, J.~Hays, P.~Perona, D.~Ramanan, P.~Doll{\'a}r, and C.~L. Zitnick, ``Microsoft coco: Common objects in context,'' in \emph{Eur. Conf. Comput. Vis.}, 2014.

\bibitem{he2016deep}
K.~He, X.~Zhang, S.~Ren, and J.~Sun, ``Deep residual learning for image recognition,'' in \emph{IEEE Conf. Comput. Vis. Pattern Recog.}, 2016.

\bibitem{kirillov2019panoptic}
A.~Kirillov, K.~He, R.~Girshick, C.~Rother, and P.~Doll{\'a}r, ``Panoptic segmentation,'' in \emph{IEEE Conf. Comput. Vis. Pattern Recog.}, 2019.

\bibitem{liu2021swin}
Z.~Liu, Y.~Lin, Y.~Cao, H.~Hu, Y.~Wei, Z.~Zhang, S.~Lin, and B.~Guo, ``Swin transformer: Hierarchical vision transformer using shifted windows,'' in \emph{Int. Conf. Comput. Vis.}, 2021.

\bibitem{deng2009imagenet}
J.~Deng, W.~Dong, R.~Socher, L.-J. Li, K.~Li, and L.~Fei-Fei, ``Imagenet: A large-scale hierarchical image database,'' in \emph{IEEE Conf. Comput. Vis. Pattern Recog.}, 2009.

\bibitem{loshchilov2017decoupled}
I.~Loshchilov and F.~Hutter, ``Decoupled weight decay regularization,'' in \emph{Int. Conf. Learn. Represent.}, 2017.

\bibitem{zhang2025v}
C.-B. Zhang, J.~Ni, Y.~Zhong, and K.~Han, ``v-clr: View-consistent learning for open-world instance segmentation,'' in \emph{IEEE Conf. Comput. Vis. Pattern Recog.}, 2025.

\bibitem{li2023mask}
F.~Li, H.~Zhang, H.~Xu, S.~Liu, L.~Zhang, L.~M. Ni, and H.-Y. Shum, ``Mask dino: Towards a unified transformer-based framework for object detection and segmentation,'' in \emph{IEEE Conf. Comput. Vis. Pattern Recog.}, 2023.

\bibitem{cheng2021mask2former}
B.~Cheng, I.~Misra, A.~G. Schwing, A.~Kirillov, and R.~Girdhar, ``Masked-attention mask transformer for universal image segmentation,'' in \emph{IEEE Conf. Comput. Vis. Pattern Recog.}, 2022.

\bibitem{li2022panoptic}
Z.~Li, W.~Wang, E.~Xie, Z.~Yu, A.~Anandkumar, J.~M. Alvarez, P.~Luo, and T.~Lu, ``Panoptic segformer: Delving deeper into panoptic segmentation with transformers,'' in \emph{IEEE Conf. Comput. Vis. Pattern Recog.}, 2022.

\end{thebibliography}
\ifCLASSOPTIONcaptionsoff
  \newpage
\fi

\appendix
\subsection{Datasets}\label{appdix:datasets}
\noindent\textbf{MS COCO 2017.}
This dataset~\cite{coco} consists of over 100,000 natural images and is widely used for tasks such as object detection, instance segmentation, and panoptic segmentation. Specifically, COCO 2017 includes 118,287 training images, 5,000 validation images, and 41,000 test images, annotated with 80 commonly occurring object classes.

\noindent\textbf{NuImages.}  
The NuImages dataset~\cite{caesar2020nuscenes} contains over 90,000 images collected for autonomous driving, with bounding box annotations for 10 common street scene classes. Specifically, the dataset includes 67,279 images for training, 14,772 images for validation, and the remaining images for testing. All images have a resolution of 1600$\times$900.

\noindent\textbf{Objects365.}  
The Objects365 dataset~\cite{shao2019objects365} is a large-scale benchmark for object detection, featuring 365 classes and approximately 2 million natural images. Specifically, it includes over 1.7 million images for training and 80,000 images for validation. The dataset contains around 30 million annotated object instances.

\subsection{Ablation Study on Instructive Self-Attention}\label{appendix:inssa}
\myPara{Impact of Configurations in Instructive Self-Attention.}
We explore different configurations of instruction tokens. First, we examine the impact of varying the number of instruction tokens, as shown in~\tabref{tab:numberins}.
The results demonstrate that performance improves with an increased number of instruction tokens.
Next, we investigate the effect of incorporating instruction tokens across different decoder layers in~\tabref{tab:layerins}.
The experimental results indicate that optimal performance is achieved when all layers utilize instruction tokens.
Instruction tokens of all layers significantly contribute to guiding object queries towards one-to-many targets.
In our work, we empirically apply instruction tokens across all six decoder layers and set their number to 10.

\begin{figure}[!t]
    \centering
    \includegraphics[width=0.8\linewidth]{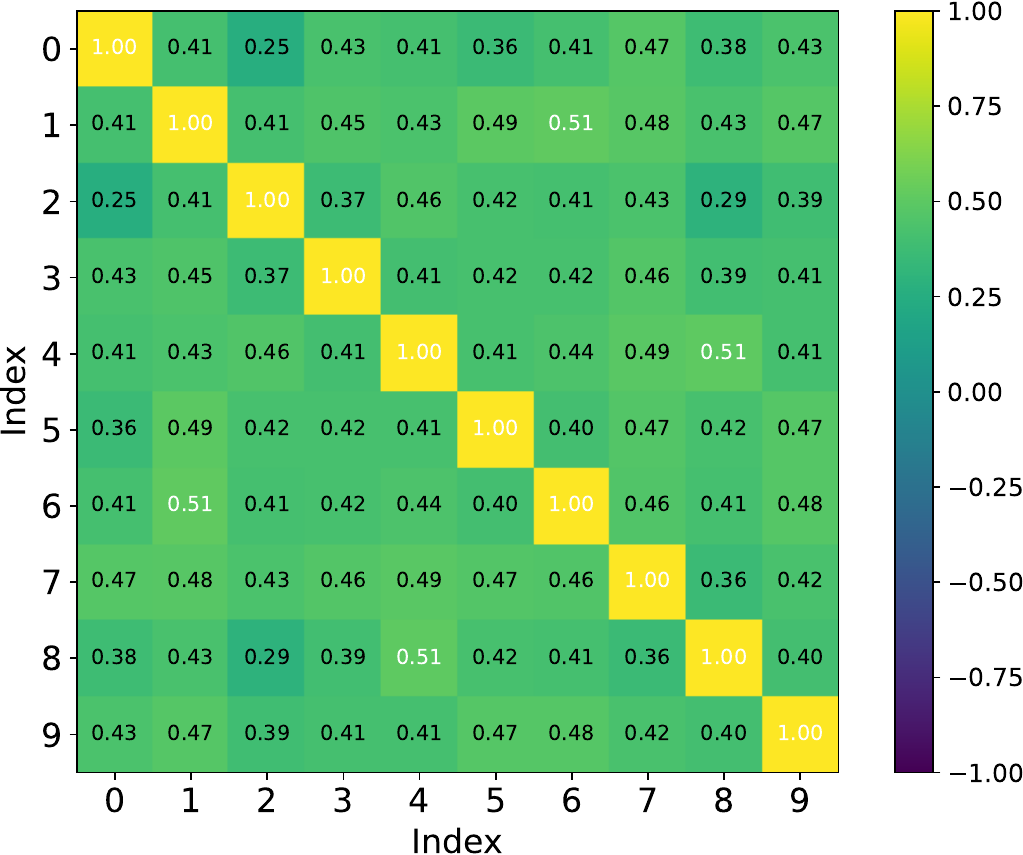}
    \caption{\textbf{The cosine similarity between 10 instruction tokens used in Mr. DETR}.}
    \label{fig:cossim}
\end{figure}

Another observation is that the model is not sensitive to the layers and numbers of instruction tokens, due to the following reasons: \textbf{(i)} Regarding the impact of layers of instruction tokens, we hypothesize that the information from instruction tokens in the first layer can be retained and utilized by subsequent layers. Meanwhile, residual connections across transformer decoder layers may help preserve the instruction information for later layers. \textbf{(ii)} To study the impact of the number of instruction tokens, as shown in~\figref{fig:cossim}, we calculate the cosine similarity between the 10 instruction tokens and find that most of them are very similar. This indicates that most instruction tokens may play a similar role, making the model insensitive to their number.

\begin{table}[h]
    \caption{\textbf{Ablation study on configurations of instruction tokens.}}
    \centering
    \begin{subtable}[!htp]{0.5\textwidth}
        \centering
        \setlength{\tabcolsep}{8pt} 
        \caption{The number of instruction tokens.} \label{tab:numberins}
        \begin{tabular}{l|ccccc}
        \toprule
        number & 1 & 5 & 10 & 50 & 100 \\
        \midrule
        AP & 50.2 & 50.1 & \textbf{50.7} & 50.4 & 50.5 \\
        \bottomrule
        \end{tabular}
    \end{subtable}%
    
    \vspace{0.05in}
    \begin{subtable}[!htp]{0.5\textwidth}
        \centering
        \caption{Layers of instruction tokens.}\label{tab:layerins}
        \begin{tabular}{l|cccccc}
        \toprule
        layers & 0 & 0$\sim$1 & 0$\sim$2 & 0$\sim$3 & 0$\sim$4 & 0$\sim$5 \\
        \midrule
        AP & 50.3 & 50.2 & 50.4 & 50.3 & 50.4 & \textbf{50.7}\\
        \bottomrule
        \end{tabular}
    \end{subtable}%
\end{table}

\subsection{Ablation Study on MoE Design}\label{appendix:moedesign}
\myPara{Impact of Configurations in MoE.}
As detailed in~\tabref{tab:configmoe}, we evaluate the effects of various MoE configurations in the encoder and decoder. Specifically, we analyze the impact of the number of experts and the top-$k$ activation strategy.
In the encoder, using 4 experts with top-1 activation yields the highest AP; however, its AP$_s$ is 0.6\% lower than that achieved with top-2 activation.
Consequently, we adopt 4 experts with top-2 activation for the encoder.
For the decoder, a configuration of 4 experts with top-2 activation delivers the best performance across the tested setups, making it our preferred choice.

\begin{table}[!t]
\caption{\textbf{Ablation study on configurations of MoE.}}\label{tab:configmoe}
    \centering
    \begin{subtable}[!htp]{0.5\textwidth}
        \centering
        \setlength{\tabcolsep}{4pt} 
        \caption{Configurations of MoE in transformer encoder} 
        \begin{tabular}{cc|cccccc}
        \toprule
        experts & top-k & AP & AP$_{50}$ & AP$_{75}$ & AP$_s$ & AP$_m$ & AP$_l$\\
        \midrule
        2 & 1 & 51.7 & 68.8 &56.5 & 35.3 & 55.3 & 66.1 \\
        4 & 1 & \textbf{51.9} & \textbf{68.9}& 56.6 & 35.0 & \textbf{55.9} & \textbf{67.2} \\
        \rowcolor{blue!10}4 & 2 & 51.8 & \textbf{68.9} & \textbf{56.7} & 35.6 & \textbf{55.9} & 67.0 \\
        4 & 3 & 51.8 & \textbf{68.9} & \textbf{56.7} & 35.0 & 55.6 & 67.0\\
        8 & 2 & 51.7 & 68.7 & 56.4& \textbf{35.8} &55.3 & 66.9 \\
        8 & 4 & 51.7 & 68.7 & 56.5 & 34.9 & 55.6 & 67.0 \\
        \bottomrule
        \end{tabular}

    \end{subtable}%
    
    \vspace{0.1in}
    \begin{subtable}[!htp]{0.5\textwidth}
        \centering
        \setlength{\tabcolsep}{4pt}
        \caption{Configurations of MoE in transformer decoder}
        \begin{tabular}{cc|cccccc}
        \toprule
        experts & top-k & AP & AP$_{50}$ & AP$_{75}$ & AP$_s$ & AP$_m$ & AP$_l$\\
        \midrule
        2 & 1 & 51.4 & 68.5 & 56.0 & 34.3 & 55.2 & 66.1\\
        4 & 1 & 51.7&68.8&56.3&35.4&55.5&66.7\\
        \rowcolor{blue!10}4 & 2 & \textbf{51.8} & \textbf{68.9} & \textbf{56.7} & \textbf{35.6} & \textbf{55.9} & \textbf{67.0} \\
        4 & 3 & \textbf{51.8} & \textbf{68.9} & \textbf{56.7} & 34.4 & 55.5 & 66.9 \\
        8 & 2 & 51.5 & 68.6 & 56.3 & 34.2 & 55.5 & 66.6\\
        8 & 4 & 51.5 & 68.6 & 56.1 & 34.1 & 55.0 & \textbf{67.0}\\
        \bottomrule
        \end{tabular}
    \end{subtable}
    
\end{table}

\begin{table}[!t]
\centering
\small
\setlength{\tabcolsep}{2pt} 
\caption{{\color{black}\textbf{Influence of factor $\phi$ used in localization-aware score calibration.}}}\label{tab:ablationscoringfactor}
\begin{tabular}{l|ccccccccccc}
\toprule
$\phi$& 0 & 0.1 & 0.2 & 0.3 & 0.4 & 0.5 & 0.6 & 0.7 & 0.8 & 0.9 & 1.0\\
\midrule 
AP & \textbf{51.8} & \textbf{51.8} & \textbf{51.8} & \textbf{51.8} & \textbf{51.8} & 51.7 & 51.7 & 51.6 & 51.6 & 51.6 & 51.5\\
AP$_{50}$ & 68.9 & 69.1 & 69.2 & 69.3 & 69.3 & 69.3& \textbf{69.4} & \textbf{69.4} & 69.3 & 69.3& 69.3\\
AP$_{75}$ & 56.7 & \textbf{56.8} & \textbf{56.8} & \textbf{56.8} & \textbf{56.8} & 56.7 & 56.7 & 56.6 & 56.6 & 56.5& 56.5\\
\bottomrule
\end{tabular}
\end{table}

\begin{figure*}[t]
  \centering
  \begin{subfigure}{0.32\textwidth}
    \includegraphics[width=\linewidth]{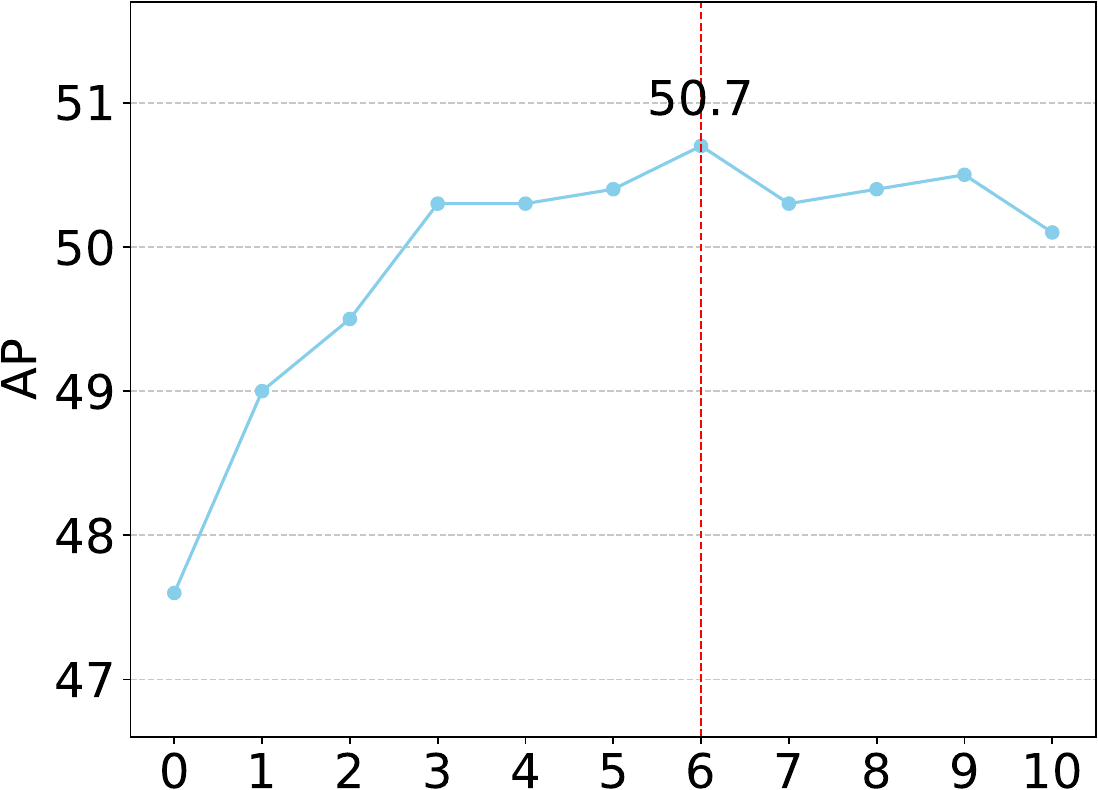}
    \caption{Ablation of $K$}\label{fig:ablationK}
  \end{subfigure}
  \begin{subfigure}{0.32\textwidth}
    \includegraphics[width=\linewidth]{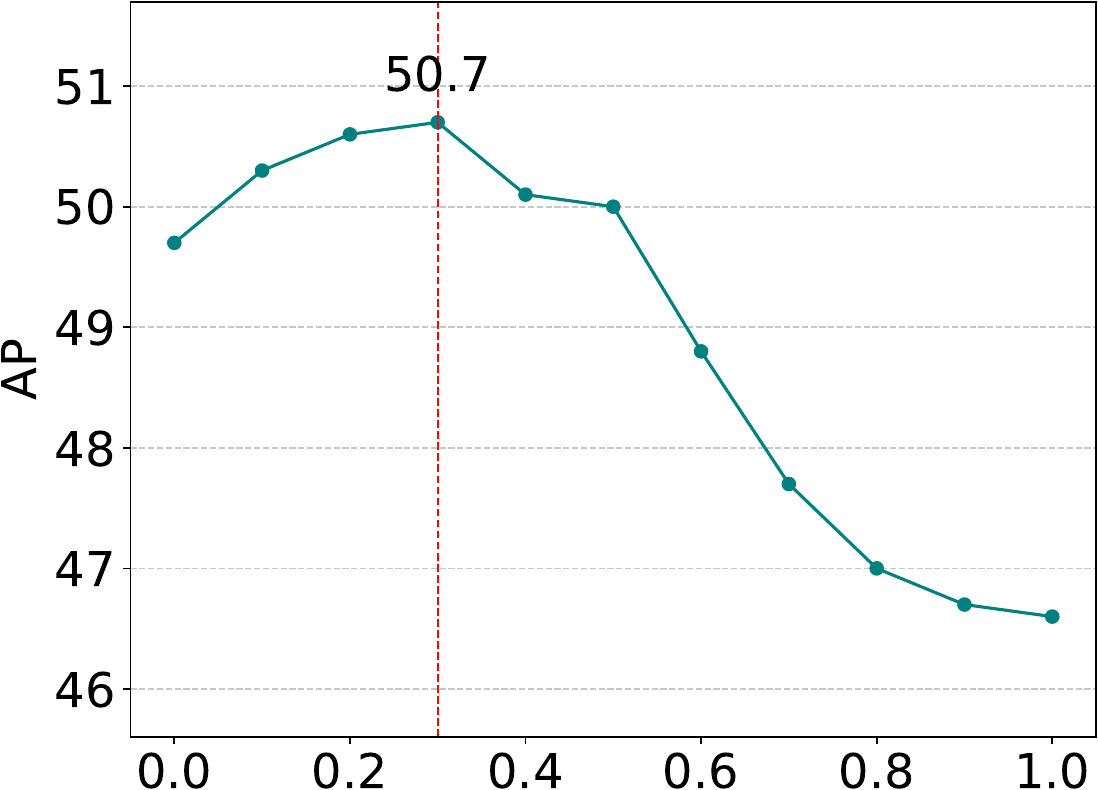}
    \caption{Ablation of $\alpha$}\label{fig:ablationalpha}
  \end{subfigure}
  \begin{subfigure}{0.32\textwidth}
    \includegraphics[width=\linewidth]{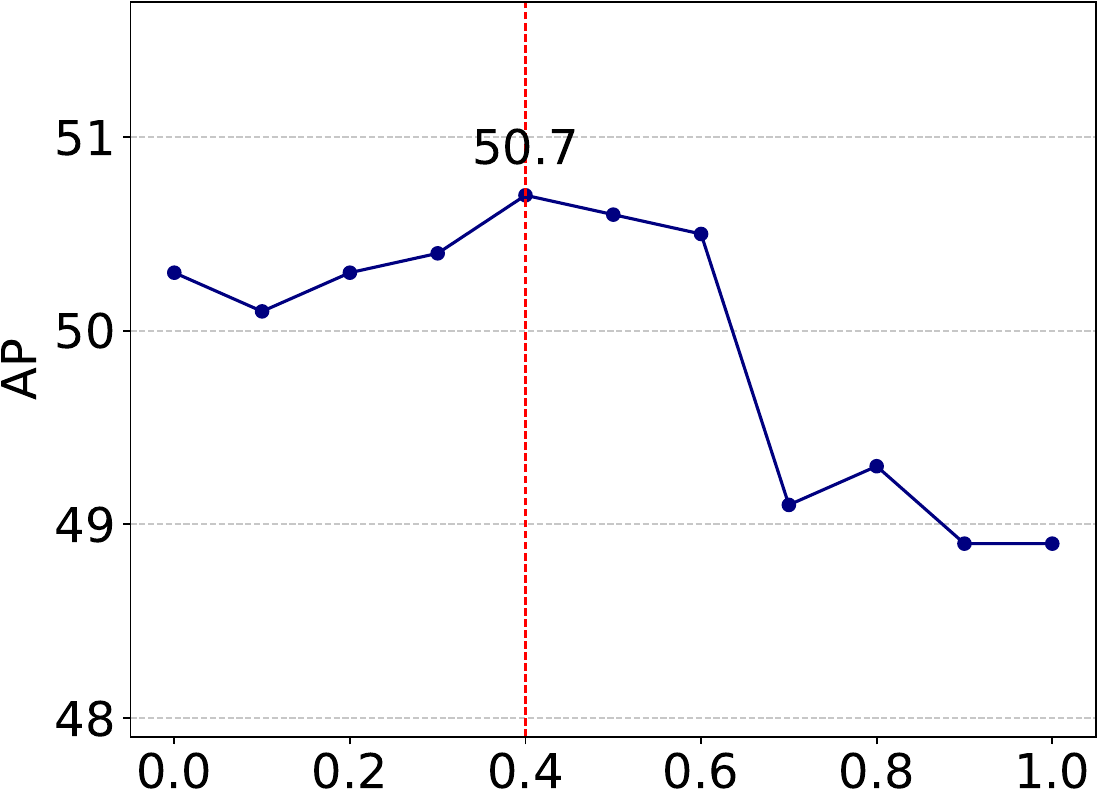}
    \caption{Ablation of $\tau$}\label{fig:ablationtau}
  \end{subfigure}
  \vskip -0.1in
  \caption{\textbf{Influence of hyper-parameters $K$, $\alpha$ and $\tau$ in the one-to-many assignment.} (a) Influence of $K$ for selecting top-$K$ positive candidates. (b) Influence of $\alpha$ that denotes the weight of classification confidence when forming the matching score $M$. (c) Influence of $\tau$ that is used to filter out low-quality candidates.} \label{fig:hyperablation}
\end{figure*}

\subsection{Ablation Study on One-to-Many Assignment}\label{appendix:assignmentshyper}
\myPara{Influence of hyper-parameters $K$, $\alpha$ and $\tau$.}
\figref{fig:hyperablation} shows the influence of the hyper-parameters $K$, $\alpha$, and $\tau$ in the one-to-many assignment~\cite{ouyang2022nms,msdetr,hu2024dac}. 
In~\figref{fig:hyperablation}(a), we observe that as the number of positive candidates increases, the model achieves its highest performance when $K = 6$. However, when $K > 7$, the one-to-many assignment increases the difficulty of removing duplicates in the primary route, leading to decreased performance. 
For the weight of classification confidence $\alpha$, as shown in~\figref{fig:hyperablation}(b), the model achieves the best performance at $\alpha=0.3$. 
Similarly, in~\figref{fig:hyperablation}(c), the performance improves as the filter threshold $\tau$ increases, reaching its peak at $\tau=0.4$.
Beyond this value, the performance declines, potentially due to the filtering out of many high-quality candidates.
In our method, we empirically set $K=6$, $\alpha=0.3$, and $\tau=0.4$.

\subsection{Ablation Study on Localization-aware Score Calibration}\label{appendix:localscore}
\myPara{Influence of $\phi$ in Localization-aware Score Calibration.}
The learned localization-aware score $s_{\text{iou}}$ is utilized to calibrate the classification score during inference.
For simplicity, $s_{\text{iou}}$ is not used during training.
To evaluate its impact, we analyze the effect of the parameter $\phi$ on classification score calibration.
As shown in~\tabref{tab:ablationscoringfactor}, increasing $\phi$ raises the weight of the classification score, leading to a slight decrease in overall AP.
Specifically, AP$_{50}$ improves, while AP$_{75}$ declines. This behavior suggests that $s_{\text{iou}}$ is more effective at distinguishing candidate boxes with IoU greater than 0.75, thereby enhancing AP$_{75}$ performance when $\phi$ is appropriately tuned.
For simplicity, we directly set $\phi$ as 0 in experiments by default.

\begin{table*}[t]
\caption{\textbf{Performance of intermediate layers of each route in terms of Box AP.} All models are trained for 24 epochs with the instance segmentation task. Route-2 is the primary route for one-to-one prediction. Route-1 and Route-3 use NMS as the post-processing.}\label{tab:intermediate}
\centering
\setlength{\tabcolsep}{9pt} 
\begin{tabular}{c|c|ccc|ccc}
\toprule
\multicolumn{1}{c|}{\multirow{2}{*}{Layer}} & \multicolumn{1}{c|}{\multirow{2}{*}{Deformable-DETR++~\cite{deformable}}} & \multicolumn{3}{c|}{Mr. DETR}                                                            & \multicolumn{3}{c}{Mr. DETR++}                                                          \\ \cline{3-8} 
\multicolumn{1}{c|}{}                       & \multicolumn{1}{c|}{}                       & \multicolumn{1}{c}{Route-1} & \multicolumn{1}{c}{Route-2} & \multicolumn{1}{c|}{Route-3} & \multicolumn{1}{c}{Route-1} & \multicolumn{1}{c}{Route-2} & \multicolumn{1}{c}{Route-3} \\ 
\midrule
0 & 41.3 & 48.2 (\textcolor{blue}{+5.6}) & 42.6 & 48.2 (\textcolor{blue}{+5.6}) & 48.7 (\textcolor{blue}{+4.9}) & 43.8 & 48.8 (\textcolor{blue}{+5.0})\\
1 & 45.1 & 49.6 (\textcolor{blue}{+2.5}) & 47.1 & 49.7 (\textcolor{blue}{+2.6}) & 49.8 (\textcolor{blue}{+1.6}) & 48.2 & 49.9 (\textcolor{blue}{+1.7})\\
2 & 47.2 & 50.1 (\textcolor{blue}{+0.9}) & 49.2 & 50.2 (\textcolor{blue}{+1.0}) & 50.3 (\textcolor{blue}{+0.2}) & 50.1 & 50.6 (\textcolor{blue}{+0.5})\\
3 & 48.1 & 50.2 (\textcolor{blue}{+0.2}) & 50.0 & 50.3 (\textcolor{blue}{+0.3}) &50.9 (\textcolor{gray}{+0.0}) & 50.9 & 51.1 (\textcolor{blue}{+0.2})\\
4 & 48.6 & 50.5 (\textcolor{blue}{+0.2}) & 50.3 & 50.4 (\textcolor{blue}{+0.1}) & 51.1 (\textcolor{gray}{+0.0}) & 51.1 & 51.2 (\textcolor{blue}{+0.1})\\
5 & 48.6 & 50.4 (\textcolor{blue}{+0.1}) & 50.3 & 50.4 (\textcolor{blue}{+0.1}) & 51.1 (\textcolor{gray}{+0.0}) & 51.1 & 51.0 (\textcolor{red}{-0.1})\\
\bottomrule
\end{tabular}

\end{table*}

\subsection{Performance of Intermediate Layers and Each Route}\label{appendix:detailedperformance}
Typically, DETR-like object detectors consist of six layers each in their transformer encoders and decoders.
Our approach for mask prediction is based on Deformable-DETR++~\cite{deformable}, utilizing solely the last decoder layer.
All six decoder layers are employed for object detection tasks.
Therefore, we only evaluate the box prediction for all layers as shown in~\tabref{tab:intermediate}.
Evaluation results suggest that Mr. DETR and Mr. DETR++ both can effectively improve the performance of the primary route across all six decoder layers, demonstrating the efficacy of our approach.
Moreover, the one-to-many prediction training routes, namely, Route-1 and Route-3, significantly surpass the primary route in the shallower layers.
For example, with Mr. DETR, Route-3 achieves a 5.6\% improvement over the primary route in layer 0, and a 0.1\% improvement in layer 5.
These experiments indicate that the primary route needs more decoder layers to reach comparable performance as the auxiliary routes equipped with NMS.

\subsection{Qualitative Results.}
\begin{figure*}[htb]
    \centering
    \includegraphics[width=1.0\linewidth]{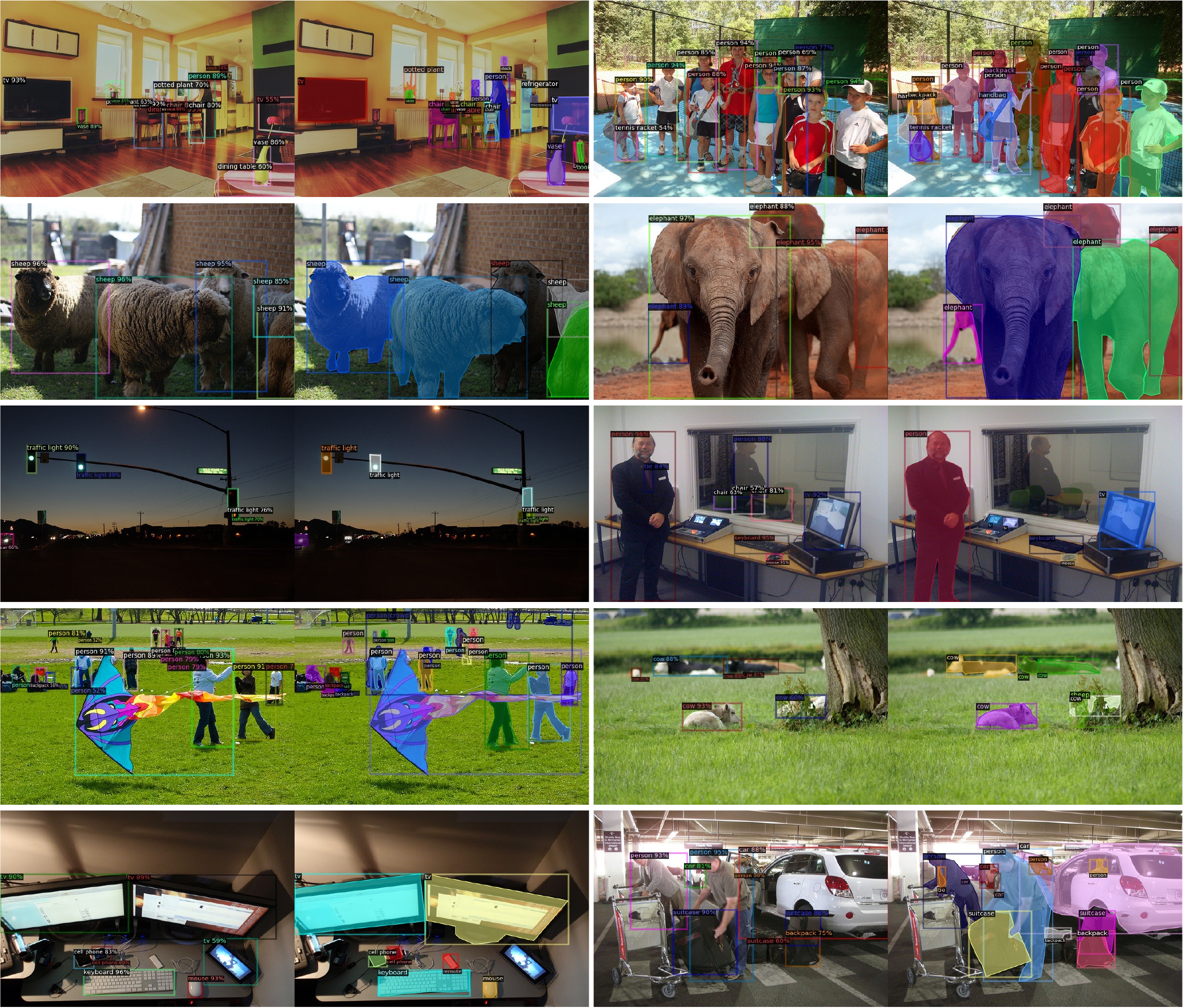}
    \caption{\textbf{Qualitative results of our method.} Left: prediction results. Right: ground truth.}
    \label{fig:visualization}
\end{figure*}

In~\figref{fig:visualization}, we present the visualization of prediction results for Mr. DETR using Swin-L~\cite{liu2021swin} as the backbone. The visualizations demonstrate that our model performs robustly across diverse scenarios, including varying object sizes and challenging low-light conditions, thereby underscoring the effectiveness of our approach.

\end{document}